\documentclass[10pt,twocolumn,letterpaper]{article}

\usepackage{cvpr}
\usepackage{times}
\usepackage{epsfig}
\usepackage{graphicx}
\usepackage{amsmath}
\usepackage{amssymb}
\usepackage{pifont}

\usepackage[table]{xcolor}
\definecolor{col1}{RGB}{232, 161, 148}
\definecolor{col2}{RGB}{148, 187, 232}
\definecolor{lightblue}{RGB}{225, 225, 255}
\definecolor{darkgreen}{RGB}{0, 150, 0}
\definecolor{darkred}{RGB}{200, 0, 0}

\definecolor{dsectioncolor}{RGB}{245, 245, 245}
\definecolor{ssectioncolor}{RGB}{175, 238, 255}%{245, 245, 255}
\definecolor{hrsectioncolor}{RGB}{255, 228, 196}
\definecolor{cropsectioncolor}{RGB}{255, 155, 155}
\definecolor{mssectioncolor}{RGB}{255, 200, 233}

\newcommand{\jw}[1]{{\color{blue}\bf [JW: #1]}}

\newcommand{\dt}[1]{{}}

\newcommand{\rev}[2]{{}{#2}}

\newcommand{\cmark}{{\color{darkgreen} \ding{51}}}%
\newcommand{\xmark}{{\color{darkred} \ding{55}}}%

\usepackage[breaklinks=true,bookmarks=false]{hyperref}

\cvprfinalcopy % *** Uncomment this line for the final submission

\def\cvprPaperID{6191} % *** Enter the CVPR Paper ID here

\ifcvprfinal\fi
\begin{document}

%%%%%%%%% TITLE
\title{Self-Supervised Monocular Depth Hints}
% \title{Depth Hints for Self-Supervised Monocular Depth Estimation}

\author{Jamie Watson$^1$
\hspace{20pt}
Michael Firman$^1$
\hspace{20pt}
Gabriel J. Brostow$^{1,2}$
\hspace{20pt}
Daniyar Turmukhambetov$^1$
\\
\hspace{-15pt}
$^1$Niantic
\hspace{20pt}
$^2$UCL\\
{\color{magenta} \url{www.github.com/nianticlabs/depth-hints}
%\vspace{-10pt}
}
}

%\author{Cl\'ement Godard$^{1}$\hspace{20pt}Oisin Mac Aodha$^{2}$\hspace{20pt}Michael Firman$^{3}$\hspace{20pt}Gabriel Brostow$^{3,1}$\\$^{1}$UCL \hspace{30pt} $^{2}$Caltech \hspace{30pt} $^{3}$Niantic\\\url{www.github.com/nianticlabs/monodepth2}}

\maketitle
%\thispagestyle{empty}

%%%%%%%%% ABSTRACT
\begin{abstract}
%Self-supervised training is a promising method for training high quality monocular depth estimators from stereo pairs or monocular videos.
Monocular depth estimators can be trained with various forms of self-supervision from binocular-stereo data to circumvent the need for high-quality laser scans or other ground-truth data. 
The disadvantage, however, is that the photometric reprojection losses used with self-supervised learning typically have multiple local minima. These plausible-looking alternatives to ground truth can restrict what a regression network learns, causing it to predict depth maps of limited quality. As one prominent example, depth discontinuities around thin structures are often incorrectly estimated by current state-of-the-art methods.

Here, we study the problem of ambiguous reprojections in depth prediction from stereo-based self-supervision, and introduce Depth Hints to alleviate their effects.
Depth Hints are complementary depth suggestions %integrated into our self-supervised training, and 
obtained from simple off-the-shelf stereo algorithms.
These hints enhance an existing photometric loss function, and are used to guide a network to learn better weights. They require no additional data, and are assumed to be right only sometimes. We show that using our Depth Hints gives a substantial boost when training several leading self-supervised-from-stereo models, not just our own. Further, combined with other good practices, we % outperform all previously proposed methods, and produces
produce state-of-the-art depth predictions on the KITTI benchmark.

%As a baseline we develop a simple model directly supervised by off-the-shelf stereo algorithms.
%Surprisingly, we find that it outperforms many recent self-supervised methods.
%However, our self-supervised Depth Hints model outperforms both the baseline and all previously proposed methods, leading to state-of-the-art results on the KITTI dataset.
\end{abstract}

%we also find that a simple model, directly supervised by off-the-shelf stereo algorithms, provides a surprisingly strong baseline which outperforms many recent self-supervised methods.

%%%%%%%%% BODY TEXT

%%%%%%%%%%%%%%%%%%%%%%%%%%%%%%%%%%%%%%%%%%%%%%%%%%%%%%
\section{Introduction}
%%%%%%%%%%%%%%%%%%%%%%%%%%%%%%%%%%%%%%%%%%%%%%%%%%%%%%

%Estimating a depth map for each single RGB image is a compelling challenge.
As the accuracy of depth-from-color algorithms improves, new opportunities are unlocked in augmented reality, robotics, and autonomous driving.
Per-pixel, ground truth depth supervision is difficult to acquire, requiring cumbersome and expensive depth-sensing devices~\cite{Geiger2012CVPR}. 
As an alternative, there is an active search for \emph{self-supervised} depth-estimation models, where a training signal is derived from data captured using commodity color cameras.
In such self-supervised settings, training involves adjusting a network's depth predictions to minimize a photometric loss. This loss is usually the distance between a reference image and the depth-guided reprojection of other views into that reference viewpoint.
%In this self-supervised setting, depth predictions are optimized at training time by minimizing the photometric distance between the reference view and depth-guided reprojection of other views into the reference viewpoint.
Depth regression is optimized and relative poses come from stereo camera calibration in a 
%Relative poses between training images are given by camera calibration in a
training-from-stereo setting \cite{godard2017unsupervised, garg2016unsupervised, mehta2018structured, poggi20183net, pillai2018superdepth}, while depth values and camera poses can be optimized jointly when training on %monocular
videos~\cite{zhou2017unsupervised, luo2018every, yang2017unsupervised, mahjourian2018unsupervised, geonet2018, wang2017learning, zou2018df, yang2018lego, ranjan2018adversarial}.

The photometric distance between the reference and depth-reprojected images could be measured with $L_1$ or $L_2$ distance, more complicated structural dissimilarity distances (DSSIM~\cite{wang2004image}), or a combination of DSSIM+$L_1$ distances~\cite{lossfunctions,godard2017unsupervised} used in state-of-the-art methods. A drawback of self-supervision is that finding the optimal depth value is normally difficult, especially where the photometric loss can be low for \emph{multiple} depth values (\eg due to repeating structures and uniformly textures areas). Consequently, training is harder, which leads to lower accuracy predictions. %that are not globally optimal.

When training depth-from-color models, our Depth Hints offer a specific alternative to the model's current depth predictions. Where the alternative's reprojection is better, the training proceeds in following the ``hint.'' Surprisingly, simply using our Depth Hints as labels for direct supervision already gives a nearly state of the art baseline. Overall, our contributions are:
\begin{enumerate}
\itemsep0em
\item We show that existing self-supervised regression methods can struggle during training to find the global optimum when minimizing photometric reprojection loss. % loss function.
\item We demonstrate that our selective training using Depth Hints is a general enhancement that can improve multiple leading self-supervised training algorithms, allowing our implementations to reach better minima. The Depth Hints can come from the same stereo image data, via, \eg OpenCV's stereo estimates~\cite{sgbm, sgbm2}. 
%\item As a baseline, we trained a neural network to regress depth that is computed from stereo using off-the-shelf stereo algorithm (OpenCV's SGM~\cite{sgbm, sgbm2}). Surprisingly, this baseline network is very competitive on the KITTI dataset~\cite{Geiger2012CVPR}, outperforming many previously published self-supervised methods trained from stereo. To the best of our knowledge, we are the first to train such a baseline model and achieve competitive results.
\item We show that our selective training with Depth Hints, coupled with sensible network design choices, leads us to outperform most other algorithms. %networks trained with just reprojection loss or supervised regression loss. \dt{needs experimental verification} 
We achieve state-of-the-art results on the KITTI dataset~\cite{Geiger2012CVPR}, outperforming both our baseline model and previously published results. % by a significant margin \gb{I think we're still happy with that statement?}.\dt{We can drop \emph{significant} from the statement. We beat Monodepth2 with \emph{sum} loss.} 
\end{enumerate}

%We enhance the established form of the photometric consistency loss when training depth-from-color models. % in the self-supervised depth estimation settings. \dt{previous sentence needs changing}
%an alternative depth value, a Depth Hint,with potentially better reprojection loss
%We note that apart from this photometric loss, where we innovate, 
%many other design decisions distinguish flavors of self-supervision. 

%We hypothesize that regardless of pre-training, online vs batch pose-estimation, cropping \vs scaling, \etc, supplying the photometric loss with Depth Hints should give more accurate depth predictions. 

%%%%%%%%%%%%%%%%%%%%%%%%%%%%%%%%%%%%%%%%%%%%%%%%%%%%%%
\section{Related Work}
%%%%%%%%%%%%%%%%%%%%%%%%%%%%%%%%%%%%%%%%%%%%%%%%%%%%%%
A neural network that predicts depth from a single image could be trained with supervised depth data, or using self-supervision by exploiting photometric consistency. The many flavors of self-supervision differ by design, opting for pre-training, cropping \vs scaling, use of synthetic data, online \vs batch pose-estimation, \etc. Here we discuss the current leading methods, and where we expect Depth Hints are and are not applicable.

%We hypothesize that regardless of pre-training, online vs batch pose-estimation, cropping \vs scaling, \etc, supplying the photometric loss with Depth Hints should give more accurate depth predictions. 

\subsection{Self-supervised depth prediction}

Self-supervised approaches can exploit photometric consistency in binocular stereo pairs, in consecutive video frames, or in consecutive frames of a stereo video. 

\textbf{Stereo training:}
Garg~\etal~\cite{garg2016unsupervised} formulated the self-supervised training of monocular depth estimation with photometric consistency loss between stereo pairs. They chose an $L_2$ loss, which tends to generate blurry results. Godard~\etal~\cite{godard2017unsupervised} (Monodepth) used a weighted sum of DSSIM \cite{wang2004image} and $L_1$ measures between correspondences. %, which was shown to produce superior results in image restoration tasks by Zhao~\etal~\cite{lossfunctions}.
They regularized network predictions with left-right consistency between left and right disparity maps and introduced a post-processing technique that boosts depth quality, where the final depth map is a weighted average of network predictions generated from the original and horizontally flipped images.
The left-right consistency was extended to a trinocular assumption by \cite{poggi20183net} for improved results.

Computing reprojection loss at a higher resolutions has been shown to improve depth map quality \cite{monodepth2, pillai2018superdepth, luo2018every}.
% in \jw{not sure I follow} Godard~\etal~\cite{monodepth2}, Pillai~\etal's SuperDepth~\cite{pillai2018superdepth}, and Luo~\etal's EPC++~\cite{luo2018every}. 
Pillai~\etal~\cite{pillai2018superdepth} also introduced differentiable flip augmentation and subpixel convolutions for increased fidelity of depth maps. Depth Hints are computed from binocular stereo data, so should be able to enhance training for any of these stereo-derived models that use the very effective DSSIM+$L_1$ photometric loss. %, and thus could be incorporated into any of these stereo-derived models.

%the photometric loss for any of these stereo-derived models.

\textbf{Monocular training:}
SfMLearner by Zhou~\etal~\cite{zhou2017unsupervised} was the first method to train a depth prediction network from monocular video only. Their network jointly predicts depth and relative camera pose changes from a frame at time $t$ to frame $t-1$, and from frame $t$ to $t+1$. Using these predictions, both the future and past frames are reprojected into the current frame, and an $L_1$ loss is applied. Additionally, this per-pixel loss is multiplied by a predicted mask to enable occluded pixels to be ignored.

\rev{}{Godard~\etal~\cite{monodepth2} build upon this, proposing that instead of averaging the loss from the reprojected future and past frames, the minimum of reprojection losses should be minimized. They also propose during training to detect and ignore pixels that appear to be stationary with respect to ego-motion.}
% Dealing with moving objects was a focus
\rev{\jw{More monocular related works, some of: luo2018every, yang2017unsupervised, mahjourian2018unsupervised, geonet2018,  wang2017learning, zou2018df, yang2018lego,\cite{ranjan2018adversarial}}}{
% Wang~\etal~\cite{wang2017learning} proposed to incorporate differentiable Direct Visual Odometry component to estimate the camera poses between multiple consecutive frames of a video. 
Multiple works propose additional regularization of predicted depths, such as surface normal consistency~\cite{yang2017unsupervised}, edge consistency~\cite{yang2018lego} and 3D pointcloud consistency~\cite{mahjourian2018unsupervised}. Recently, multiple works % have focused on jointly modeling the depth, motion of objects and the motion of the camera. For example, 
~\cite{luo2018every, ranjan2018adversarial, geonet2018, zou2018df} have proposed to model the relationship of pixels in the consecutive frames of a video with joint estimation of optical flow, depth and camera poses with loss terms that supervise the different estimates to be consistent.
Depth Hints are not naturally compatible with monocular-video only data; extensions are left as future work.
}
\dt{I left the previous sentence.}

It is also possible to train from both monocular video (forward and backward in time) and stereo pairs for improved pose and depth estimation~\cite{monodepth2, zhanst2018}.

% \jw{as per Gabe's comment, and for space saving shall we remove the below?}\dt{R1 asked to explain how Klodt et al is different from Depth Hints. We wrote about that in the rebuttal, if the paragraph stays, the text needs to emphasize the difference.}
% Depth Hints are not naturally compatible with monocular-video data. However, one could imagine two possible extensions to our method: 1) One could use a monocular-SLAM system to compute per-frame depths to serve as Depth Hints. \jw{say similar to Klodt} 2) For videos without ego-motion, the Depth Hints concept could be re-conceived for self-supervised learning of optical flow, akin to~\cite{guidedflow17}, as a kind of Flow Hint. \gb{Are we not interested in writing a Flow Hints paper? Could cut this 2nd opportunity from the text for strategic advantage. Or keep it, and add it to the patent.}\dt{This was in the original draft. We could remove the whole paragraph.}
% While we focus on self-supervised monocular depth estimation problem, we note that a similar technique could be applied in other self-supervised settings, e.g. optical flow. Cite Guided Optical Flow Learning \cite{DBLP:journals/corr/ZhuLNH17}

\subsection{Additional supervision}
Following the work of Eigen~\etal~\cite{eigen2014depth}, many others have trained using forms of per-pixel ground-truth depth labels. Training with ground truth is almost always a good idea when it is available, and we strive to push self-supervised performance closer to this ceiling. 

\textbf{With LiDAR Depth:}
Kuznietsov~\etal~\cite{kuznietsov2017semi} optimize a fused loss, which sums a supervised loss based on sparse LiDAR pointclouds and a self-supervised loss from stereo images. They follow Godard~\etal~\cite{godard2017unsupervised} by using DSSIM+$L_1$ as the photometric reprojection loss, and they follow Laina~\etal~\cite{laina2016deeper} by using berHu loss (inverse Huber)~\cite{laina2016deeper} on the LiDAR pointcloud.

Fu~\etal~\cite{fu2018deep} showed that framing the regression of depths as ordinal classification can bring significant improvements to supervised prediction, though this concept is difficult to adopt for self-supervised training.

\textbf{With Synthetic Depth:}
Synthetic data is an interesting source of ground-truth depths and/or stereo pairs.
Instead of the usual photometric loss, domain adaptation is possible using generative adversarial networks~\cite{mehta2018structured}\rev{}{, or by leveraging the ability of stereo matching networks to better generalize to real world data~\cite{guo2018learning}}.
Luo~\etal~\cite{singlestereo2018} demonstrate how synthetic data can be incorporated into single-image depth estimation with a two stage process. First, a network synthesizes a right view from the left view. Then, a second network performs stereo matching to recover depth from the half-synthetic stereo pair. Both networks can be trained on stereo+synthetic data, and optionally fine-tuned with ground truth.

\textbf{With SLAM Depth:}
Yang~\etal~\cite{yang2018deep} train a monocular depth estimation network with both self-supervision from stereo pairs, and supervision from sparse depths estimated in batch by the Stereo DSO~\cite{Wang_2017_ICCV} algorithm. They demonstrate that a depth estimator network can improve visual odometry for monocular videos, resolving some scale ambiguity. % Interestingly, the inverse depth estimates from Stereo DSO are incorporated into the objective function with a berHu loss on the \emph{disparities} rather than \emph{depth} as originally proposed by Laina~\etal~\cite{laina2016deeper}.

Klodt and Vedaldi~\cite{klodt2018supervising} use sparse depths and poses from a traditional SLAM system as a supervisory signal to train depth and pose prediction networks. They train from monocular videos (in contrast to ~\cite{yang2018deep}), which requires special consideration of scale, and modeling of uncertainty in the depth and poses.

\textbf{With Semantic Labels:}
Ramirez~\etal~\cite{ramirez2018} show that a depth estimation network can be improved by jointly predicting depth and semantic labels. They propose % using a shared encoder with separate decoders, and introduce 
a novel cross domain discontinuity loss to help align depth discontinuities with semantic boundaries.

\textbf{With Estimated Depth:}
The concurrent work monoResMatch by Tosi~\etal~\cite{Tosi_2019_CVPR} also exploits proxy ground truth labels generated with a traditional stereo matching method~\cite{sgbm}. % They train with additional berHu loss~\cite{laina2016deeper} on left-right consistent depth estimates. 
The inclusion of the proxy supervision is shown to greatly improve accuracy over using a standard self-supervised loss. Our proposed loss is different from theirs. \dt{Added the last sentence here.}

% \subsection{Non-convex optimization} \mf{We could scrap this section.} \dt{I vote for scrapping. it seems disjoint from the rest of the narrative.}
% Instead of modifying the loss as we're proposing, one could imagine using non-convex optimization to mitigate for a network learning the wrong regression function. However, optimizing non-convex losses is difficult, especially with batch training commonly used when training deep models on large datasets \cite{krizhevsky2012imagenet}.
% One method to help discourage a network from making predictions at local minima is to train an ensemble; randomization in seeds and augmentation can then lead each member to a different minimum.
% Snapshot Ensembles \cite{huang2017snapshot} are an approximation trick to help to make this process less computationally expensive, while Haeffele and Vidal~\cite{haeffele2017global} work through the pre-conditions needed for global optimality to be found. 
% In our problem, the loss isn't convex, and the networks are substantial in size.

%, but they focus on the network optimization, rather than the loss. 

%Also their paper is very theoretical and only show results on tiny nets.

% Not exactly sure what's being said here:
% https://arxiv.org/pdf/1806.09777.pdf
% but it seems to suggest that dropout can help find global minima

%%%%%%%%%%%%%%%%%%%%%%%%%%%%%%%%%%%%%%%%%%%%%%%%%%%%%%
\section{Background}
%%%%%%%%%%%%%%%%%%%%%%%%%%%%%%%%%%%%%%%%%%%%%%%%%%%%%%

In monocular depth estimation, the task is to train a neural network to predict a depth map $d$ from a single input image $I$. 
In the self-supervised setting, the training data consists of pairs of images $I$ and $I^\dagger$ with known camera intrinsics $K$ and $K^\dagger$, and relative camera pose $(R, t)$. The network is trained to reconstruct the reference image $I$ by reprojecting the other image into the reference view, so
\begin{equation}
\tilde{I} = \pi(I^\dagger, K^\dagger, R, t, K, d).
\end{equation}

Hence, pixel $i$ at predicted depth $d_i$ gets a color value $\tilde{I}_i$. Under idealized training conditions, the predicted color $\tilde{I}_i$ would perfectly match $I_i$ for all $i$.

When training from stereo, the only unknown parameter in $\pi()$ is the estimated depth $d$. For monocular or stereo video, in addition to $d$, the network also needs to predict the camera pose $(R, t)$. Presently, we do not pursue hints for pose, though this is a natural extension of our method.

Many leading algorithms now use a differentiable photometric consistency loss to measure how well the warped image approximates the reference image. We focus on the DSSIM+$L_1$ loss, a photometric consistency loss used in many self-supervised monocular depth estimation methods \cite{godard2017unsupervised, pillai2018superdepth, yang2018deep, luo2018every}. This loss is computed per pixel as
\begin{align} \label{eq:ReprojLoss}
l_r(d_i) = \alpha \frac{1 - \text{SSIM}(I_i, \tilde{I}_i)}{2} + (1 - \alpha) | I_i - \tilde{I}_i| ,
\end{align}
where SSIM$()$ is computed over a 3x3 pixel window, with $\alpha$ set to 0.85. 
% where the per-image loss $L_r = \sum^N_{i=0} l_r(d_i) / N$, and $N$ is the number of pixels.  %\jw{leave 0.85 and 0.15 as hyperparams etc?}\dt{leave as 0.85 for simplicity. we don't currently experiment with other weights. these weights are practically everywhery. DVSO uses 0.84/0.16. Superdepth uses 0.1/0.9 for pose network only.}

If we were training \emph{with} supervision, we would minimize the distance between continuous depth $d_i$ predicted by the network at pixel $i$, and depth $d'_i$ procured by a LiDAR system, Kinect sensor, a stereo algorithm, or a SLAM system, depending on the training context. Note that the last two contexts could count as a form of self-supervision, in that the labels $d'_i$ are inferred, and not ground-truth measurements. There are several supervised losses $l_s$ used and compared in the literature~\eg~\cite{laina2016deeper, eigen2014depth, Hu2018Revisiting}, such as $L_1$, $L_2$ and (names in superscripts): \vspace{-5pt}
\begin{align}
%l_{s}^{L_2}(d_i, d'_i) &= (d_i - d'_i)^2; \\
%l_{s}^{L_1}(d_i, d'_i) &= |d_i - d'_i|; \\
l_{s}^{\log L_1}(d_i, d'_i) &= \log(1 + |d_i - d'_i|); \\
l_{s}^{\text{berHu}}(d_i, d'_i) &= \begin{cases}
    | d_i - d'_i | , & \text{if $ |d_i - d'_i| \leq \delta $},\\
    \frac{(d_i - d'_i)^2 + \delta ^2}{2 \delta}, & \text{otherwise}.
  \end{cases}
\end{align}  
Typically $\delta = 0.2 \max_{i=0..N}(|d_i - d'_i|)$. % Depending on the reliability of the provided depth $d'_i$ you may choose one of the proposed depth values.
Similarly, the same losses are often applied on inverse depth (\ie disparity).
We found that $l_{s}^{\log L_1}$ works well %best \jw{didnt actually try huber - do we just mentioned in the next section that we chose this and leave it at that?}\dt{good point. need to say that we chose it for some arbitrary reason.} 
with estimated depths (and \cite{Hu2018Revisiting} favors it for Kinect data), while $l_{s}^{\text{berHu}}$ is an established choice for accurate LiDAR and SLAM depths~\cite{kuznietsov2017semi, laina2016deeper} and disparities~\cite{yang2018deep}.

% %\dt{we should remove respective, as \cite{laina2016deeper} compares $L_1$, $L_2$ and berHU} and defined as\mf{ interesting enough we could cite papers which use each loss}
% \begin{equation}
% l_{s}^{L_2}(d_i, d'_i) = (d_i - d'_i)^2;
% \end{equation}
% \begin{equation}
% l_{s}^{L_1}(d_i, d'_i) = |d_i - d'_i|;
% \end{equation}
% \begin{equation}
% l_{s}^{\log L_1}(d_i, d'_i) = \log(1 + |d_i - d'_i|);
% \end{equation}
% \begin{equation}
% l_{s}^{\text{berHu}}(d_i, d'_i) = \begin{cases}
%     | d_i - d'_i | , & \text{if $ |d_i - d'_i| \leq \delta $},\\
%     \frac{(d_i - d'_i)^2 + \delta ^2}{2 \delta}, & \text{otherwise}.
%   \end{cases}
% \end{equation}

\begin{figure*}[t]
\begin{minipage}[c]{\linewidth}
\begin{minipage}[c]{\linewidth} % 4 rows of minipages: 3, 3, 7, 7
\centering
\begin{minipage}[c]{0.99\linewidth} % 3 big figures
\centering
\begin{tabular}
{
*{3}{p{0.3\linewidth}}
}
\centering\includegraphics[width=\linewidth]{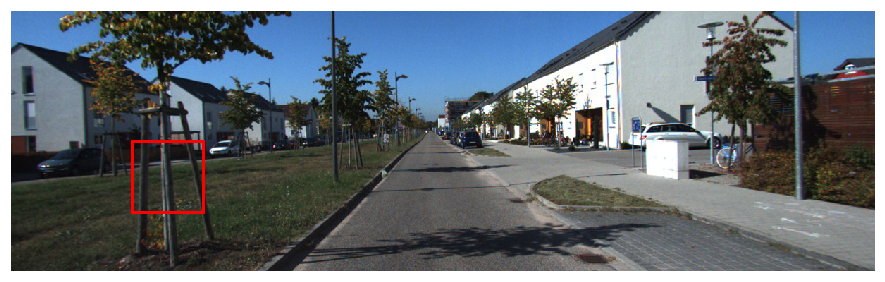}&
\centering\includegraphics[width=\linewidth]{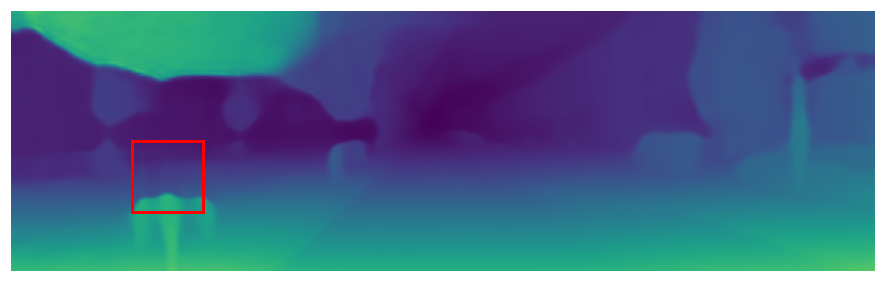}&
\centering\includegraphics[width=\linewidth]{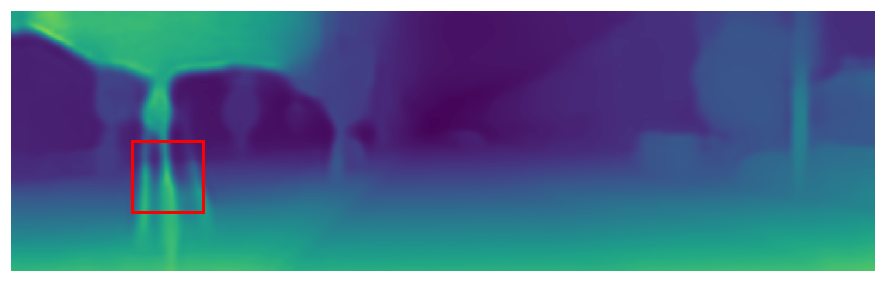}
\end{tabular}
\end{minipage}
\\
\vspace{-3pt}
\begin{minipage}[c]{0.99\linewidth} % 3 captions
{\small
\begin{tabular}
{
*{3}{p{0.3\linewidth}}
}
\centering Training Image &
\centering Without Depth Hints &
\centering With Depth Hints
\end{tabular}}
\end{minipage}
\\
\vspace{-3pt}
\begin{minipage}[c]{0.99\linewidth} % 7 small figures
\centering
\bgroup
\setlength{\tabcolsep}{-3pt}
\begin{tabular}
{
*{6}{p{0.17\linewidth}}p{0.043\linewidth}
}
\centering\includegraphics[width=\linewidth]{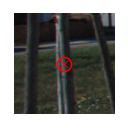}&
\centering\includegraphics[width=\linewidth]{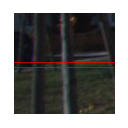}&
\centering\includegraphics[width=\linewidth]{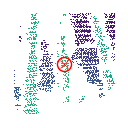}&
\centering\includegraphics[width=\linewidth]{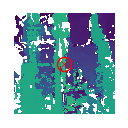}&
\centering\includegraphics[width=\linewidth]{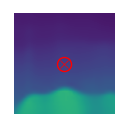}&
\centering\includegraphics[width=\linewidth]{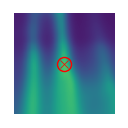}&
\centering\includegraphics[width=\linewidth]{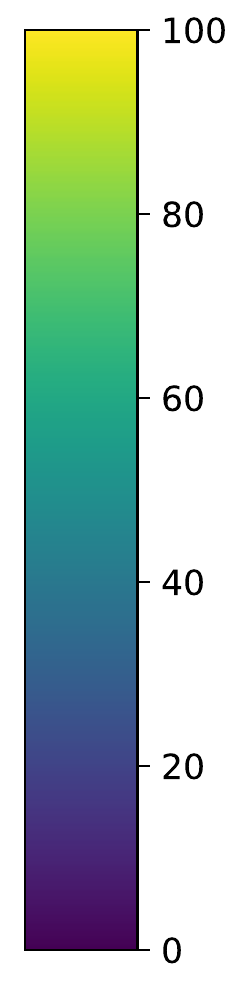}
\end{tabular}
\egroup
\end{minipage}
\\
\vspace{-5pt}
\begin{minipage}[c]{0.99\linewidth} % 7 captions
{\small
\bgroup
\setlength{\tabcolsep}{-3pt}
\begin{tabular}
{
*{6}{p{0.17\linewidth}}p{0.043\linewidth}
}
\centering Image Patch &
\centering Other View &
\centering LiDAR &
\centering Fused SGM &
\centering Without Depth Hints &
\centering With Depth Hints &
\centering Colormap
\end{tabular}
\egroup
}
\end{minipage}
\end{minipage}
\\
\begin{minipage}[c]{0.99\linewidth}
\centering
\includegraphics[width=0.58\linewidth]{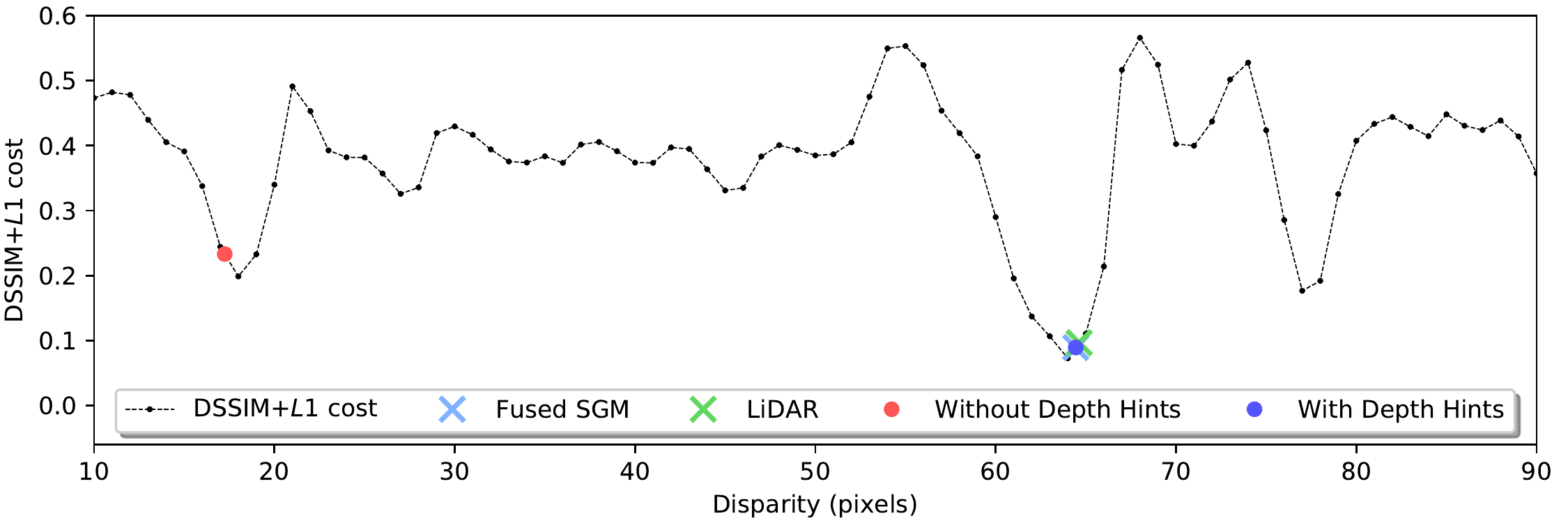}
\includegraphics[width=0.195\linewidth]{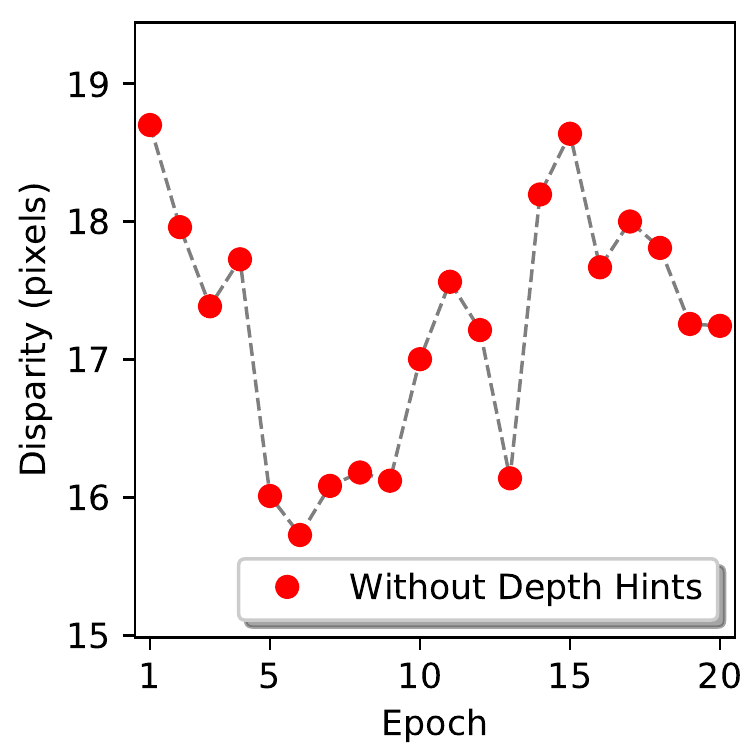}
\includegraphics[width=0.195\linewidth]{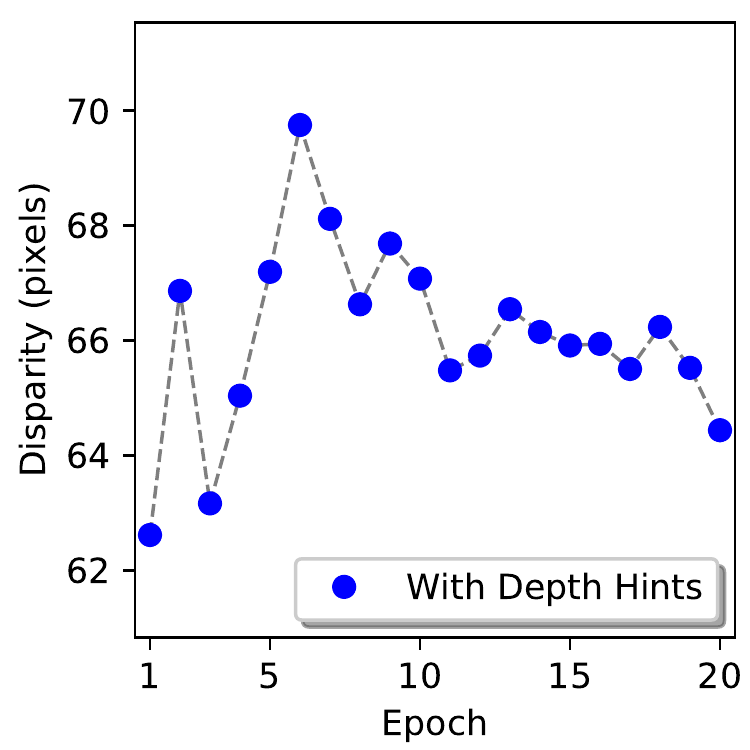}
\end{minipage}

\end{minipage}
\caption{\textbf{Top row:} Image from the training set and corresponding depth maps produced by neural networks trained without and with Depth Hints (Godard~\etal~\cite{monodepth2} (Monodepth2) architecture and loss).
\textbf{Middle row, left to right:} Crop of the image centered around a thin structure with the center pixel circled, the scanline in the other image for the circled pixel, LiDAR pointcloud, fused depth map from SGM, crop of the depth map produced by a network trained without Depth Hints, our result, and the color coding illustrating pixel disparities.
% \textbf{Bottom row:} Plot of DSSIM+$L_1$ cost of the pixel on the thin structure for every pixel disparity. Line segments in red and blue show the photometric loss on network predictions for the centered pixel while training. The number $q$ corresponds to the prediction made by the network after $q$ epochs. The network trained {\color{red} without Depth Hints} gets stuck in a local minimum and does not escape even after 20 epochs. On the other hand, the network trained {\color{blue} with Depth Hints} is in the vicinity of the correct solution after the first epoch.
\textbf{Bottom row:} On the left is the plot of DSSIM+$L_1$ cost of the pixel on the thin structure for every pixel disparity. Plots on the right show the predictions made by the network after $q$ epochs when trained with and without Depth Hints. The network trained {\color{red} without Depth Hints} gets stuck in a local minimum and does not escape even after 20 epochs. On the other hand, the network trained {\color{blue} with Depth Hints} is in the vicinity of the correct solution (disparity of 64.63 according to LiDAR) after the first epoch.
We visualize depths as disparities in pixel space for clarity. (Best viewed in color.)
}
\vspace{-13pt}
\label{fig:motivation}
\end{figure*}

%%%%%%%%%%%%%%%%%%%%%%%%%%%%%%%%%%%%%%%%%%%%%%%%%%%%%%
\section{The Need for Depth Hints}
%%%%%%%%%%%%%%%%%%%%%%%%%%%%%%%%%%%%%%%%%%%%%%%%%%%%%%
% \setlength{\tabcolsep}{2pt}

Figure~\ref{fig:motivation}~(top) shows an input image from the training set, and the corresponding depth map produced by Godard~\etal~\cite{monodepth2}'s network, trained on stereo data with DSSIM+$L_1$ loss. We can see that the network failed to converge to the correct solution, with many thin structures missing in the predicted depth map.

How do these mistakes come about? It is not failure to generalize or the result of overfitting, as this is an image from the training set. Another explanation could be that the depth map's artifacts are due to a poor choice of photometric reprojection loss, where failures on thin structures are not penalized enough. However, Figure~\ref{fig:motivation}~(bottom) shows DSSIM+$L_1$ loss for a pixel on a thin object, and we can see that the loss is lower still for more appropriate depth values.

We hypothesize that, in the absence of a ground-truth depth label, the network becomes stuck, learning to regress depth for a local minimum of the reprojection loss and failing to seek the global minimum. To escape such bad minima, we propose to consult an alternative depth value in case it can offer a more plausible reprojection, and if so, incorporate it into the objective function. We refer to these alternative depth values as Depth Hints. Depth Hints, born from noisy estimates, can be more or less accurate than our current network prediction, and therefore we expect the iterative training of a CNN to gradually change its uptake of these hints as it converges.
In contrast to supervised depth prediction, though, our main focus is to converge to the best minimum using a standard self-supervised reprojection loss. Depth Hints are only used, when needed, to guide the network out of local minima.

%%%%%%%%%%%%%%%%%%%%%%%%%%%%%%%%%%%%%%%%%%%%%%%%%%%%%%
\section{Method}
%%%%%%%%%%%%%%%%%%%%%%%%%%%%%%%%%%%%%%%%%%%%%%%%%%%%%%
We assume that stereo data is being used to train a CNN to regress a depth map from a color image. We start from an existing loss function, designed for self-supervised training from such stereo images, that uses a photometric reprojection measure like DSSIM+$L_1$.
%We assume that an existing loss function, designed for self-supervised training from stereo images, uses a photometric reprojection measure like DSSIM+$L_1$ to learn to regress depth maps. 
We propose to adaptively modify the existing training process only where the currently estimated depth map is worse than the Depth Hint. %, an alternative depth map. \dt{Do we want to mention that this is similar to pretraining?} % \jw{actually since we use a threshold of 0 we ignore only when Depth Hints are worse}
A Depth Hint is essentially a depth map estimated by a third-party binocular stereo algorithm.
%We could choose to get Depth Hints by various means. For clarity, in our learn-from-stereo self-supervised setting, we use a standard heuristic-based stereo algorithm (OpenCV's SGM~\cite{sgbm, sgbm2}), instead of stronger learning based stereo algorithms, \eg \cite{cheng2018learning, chang2018pyramid, song2019stereo},  that require still further ground truth depths for training.

\subsection{Training from stereo pairs} \label{sec:TrainingFromDHstereo}
During training, we provide our network with a per-pixel Depth Hint, \ie a potential alternative hypothesis to the network's own depth estimate. Our key idea is that we only want to provide a supervisory signal from the Depth Hints in places where they make for a superior reprojected image $\tilde{I}$, compared to using the network prediction. Else the hint is ignored. %If the network prediction computes a loss value that is better than the loss at the proposed Depth Hint, then the Depth Hint is ignored. 
To be clear, the proposed objective is not learning to regress a map of hinted depth values. That would be a supervised loss, and is indeed one of our baselines. Interestingly, \cite{garg2016unsupervised} explored that baseline and found it disappointing, because $L_2$ was in favor at the time. Rather, our objective remains to optimize a given algorithm's existing loss, % \gb{or existing loss? Point is, don't we show Depth Hints helping even when the loss isn't DSSIM+$L_1$?}\dt{existing and reprojection are basically the same in this context. happy to change it with existing. reprojection loss is not only DSSIM+$L_1$}, 
and to consult a pixel's Depth Hint only when the reprojection loss can be improved upon.

In light of this, we reformulate our loss for pixel $i$ as:
\vspace{-5pt}
\begin{align}\label{eq:ours}
l_{ours}(d_i) &= 
    \begin{cases}
       l_{r}(d_i) + l_{s}^{\log L_1}(d_i, h_i) & \text{if } l_{r}(h_i) < l_{r}(d_i)\\
       l_{r}(d_i) & \text{otherwise},
    \end{cases}
\end{align}
for an inferred network depth $d_i$ and a depth hint $h_i$, with an associated self-supervised loss function $l_r$ from (\ref{eq:ReprojLoss}) judging the photometric quality of the depth estimate.

%Given a current network estimate $d_i$ and a proposal (from OpenCV or elsewhere) $p_i$ the per pixel loss is:
%\begin{align}
%L_i &= 
%    \begin{cases}
%       \log(|p_i - d_i| + \epsilon ) & \text{if } c(p_i) < c(d_i)\\
%       c(d_i) & \text{otherwise,}
%    \end{cases}
%\end{align}
%where $\epsilon$ is a small positive number.
%Or:
%\begin{align}
%\eta &= [ \, c(p_i) < c(d_i) \, ] \\
%L_i &= \eta \, | p_i - d_i| + (1 - \eta) \, c(d_i),
%\end{align}
%where $[\,]$ is the Iverson bracket.

% The DSSIM+$L_1$ reprojection loss $l_{r}$ that we use throughout this paper is computed over a 3x3 window, which means that the depth hints need to be consistent in a 3x3 window around each pixel. \dt{Hence, a smooth depth hint map is preferred?} \gb{I can read this multiple ways. What is the real intent of this paragraph?} \jw{similar - when choosing whether to use a depth hint we look at SSIM over a 3x3 window, is that what we mean?} \dt{if the depth hint is a random white noise, then it will be useless, as SSIM uses 3x3 window, and 3x3 windows of white noise are not consistent as depth hypotheses.} \mf{Is this an important point? Does it lead to any design decisions we make?}

\vspace{5pt}
\noindent{\textbf{Computing Depth Hints:}} We propose to generate Depth Hints using stereo pairs. Depth Hints with perfect accuracy are unattainable, and it would be extremely expensive to sweep discrete per-pixel depth values to find those that generate the optimal DSSIM+$L_1$ reprojection. 
%\dt{Ideally, we could find the actual global minimum for DSSIM+$L_1$ reprojection loss by computing the loss at each pixel for discrete (but dense) set of depth values and retain the depth value with the best loss value. This is expensive, and }. 
Instead, we use a standard heuristically-designed stereo method to compute depth. %narrow down the search space. Since Depth Hints could be ignored by a good network prediction we would benefit from generating multiple Depth Hints for each input image. % We would also benefit if the stereo algorithm is fast, as it can then be done on the fly during training.
It is tempting to use a state-of-the-art stereo algorithm instead, \eg \cite{cheng2018learning, chang2018pyramid, song2019stereo}, %, to provide Depth Hints for our network.
however most modern stereo algorithms are supervised using the LiDAR ground truth from the KITTI dataset. Using one of these would cause us to be implicitly learning from laser-scanned ground-truth data.
Further, generating multiple depth maps is not trivial with most stereo methods.

%a `uniqueness ratio' which controls how likely it is that the algorithm will discard a match as being unreliable, and 
Semi-Global Matching (SGM)~\cite{sgbm, sgbm2} is an off-the-shelf stereo matching algorithm available in OpenCV. 
SGM allows generation of different depth maps, depending on the hyperparameters used.
For example, one can specify the size of the block to match between images, and the number of discrete disparities to evaluate. 
Hence, at training time, we can randomly choose hyperparameters for SGM to generate Depth Hints on the fly. We refer to such Depth Hints as ``Random SGM.'' Alternatively, for each training image pair, we can generate a collection of depth maps by running SGM with every possible hyperparameter choice. We discretize this space into $12$ parameter choices, formed of combinations of three block sizes with four resolutions of disparities. We call this version of Depth Hints ``Fused SGM,'' because it checks that collection of depth maps and chooses the depth value at each pixel based on the DSSIM+$L_1$ score. Fused SGM Depth Hints are pre-computed just once for the training corpus. % \jw{clarify that we use Fused SGM as "Ours"?}
Unless specified, we use Fused SGM depths as hints in our models.

Finally, SGM's depth maps can contain holes where the matching cost is ambiguous. All losses associated with SGM's depth maps are set to infinity for such pixels.

\subsection{Training from stereo video}

We can also apply this same method in the stereo \emph{video} self-supervised task, where training data is a video of binocular pairs. In addition to the depth prediction for the current frame at time $t$, the network also produces two camera poses for the forward $t+1$ and backward $t-1$ frames. The input to the depth prediction network is just the current frame $t$, while the pose prediction network is given 3 frames at times $t$, $t-1$ and $t+1$. Similarly to Godard~\etal~\cite{monodepth2}, we warp all three other views (other image of the stereo pair, forward frame and backward frame) into the reference viewpoint, and select the photometric reprojection loss as the minimum of the 3 associated losses at each pixel.

% Assume we have poses and depths from sfm or similar. 
% Pose network loss and depth network loss are both trained with our method.
% Scale invarience? (Cite the supervising old with new - rescaled to make medians the same).
% \begin{align}
% T &= \begin{cases}
%       P & \text{if }  \\
%       c(d_i) & \text{otherwise,}
%     \end{cases}
% \end{align}
% Scale-invariant pose translation loss?:
% \begin{align}
%     L &= \left\lVert 
%         \frac{T}{\lVert T \rVert_2} - \frac{\hat{T}}{\lVert \hat{T} \rVert_2} 
%     \right\rVert_1
% \end{align}
% Depth can also be trained to be scale invariant, using the Eigen loss? Or use ratios of lengths of vectors?

\subsection{Implementation Details}
%  We use PyTorch for our experiments. 
%  The Adam optimizer \cite{kingma2014adam} is used for training for $20$ epochs, except where noted. 
%  The learning rate starts at $10^{-4}$ and decreases by a factor of $10$ every $5$ epochs. 
 Our network architecture and training regime closely follow Godard~\etal~\cite{monodepth2}, and can be viewed in the supplementary materials. Unless otherwise specified, we use Resnet-18~\cite{he2016deep} as the encoder, pretrained on ImageNet~\cite{imagenet}, also following \cite{monodepth2}. We specify the resolution of the input images explicitly, as it was shown to impact accuracy~\cite{monodepth2, pillai2018superdepth}.
 
 Depth map post-processing \cite{godard2017unsupervised} improves the quality of the final depth maps, so, for the quantitative results in Tables~\ref{tab:ablation},~\ref{tab:existing} and~\ref{tab:kitti_eigen}, we add a ``PP'' column to indicate if post-processing was applied.

% \jw{Can probably remove the below as we aren't including cropped version of our model?} \gb{No crop results at all? If crop results appear only in the supplemental, or not at all, then cut (or move) from here. Probably good to use the space for extra qualitative images.}
\rev{
 Due to GPU memory restrictions, networks usually cannot be trained at full resolution, so the majority of methods train on downsampled images. However, it is possible to train on crops, where the network is fed a random crop of the full resolution image \eg as in DORN~\cite{fu2018deep}. At test time, the full resolution image is tiled into suitable crops, then each crop is processed by the network and the depth maps are averaged to produce the full resolution output. Training on crops has the potential to improve most models, because the network processes more data and is able to `see' finer details. %due to random crops, and the final result is an average of multiple predictions. 
 We specify in our results if a network was trained with crops instead of downsampling.
}{
Due to GPU memory restrictions, some methods train the network with a random crop of the full resolution image \eg as in DORN~\cite{fu2018deep}. At test time, the full resolution image is tiled into suitable crops, then each crop is processed by the network and the depth maps are averaged to produce the full resolution output. Training on crops has the potential to improve most models, because the network processes more data and is able to `see' finer details. %due to random crops, and the final result is an average of multiple predictions. 
We specify if a network was trained with crops instead of downsampling. % \jw{Because of this, we consider training on crops to be equivalent to training on high resolution images?}
}
% \dt{I'm keeping it here just to explain crops results in table 3.}
% \textbf{Network architectures}
% \mf{Check if we actually do this at the end... If so, the notation needs updating.}
% For pose prediction we follow \cite{yang2018lego, pillai2018superdepth} in using a pose network which takes as input a stack of frames $\{ I_{t-1}, I_t, I_{t+1} \}$ as input, and predicts relative camera transformations $\{ T_{t-1 \rightarrow t}, T_{t \rightarrow t+1} \}$.
% We use a ResNet18 architecture for this purpose, modified to accept three input frames (\ie nine channels) as input.

\begin{table*}[th]
  \centering
    % \begin{minipage}[l]{1.0\textwidth}
%   \resizebox{0.7\textwidth}{!}{
  \centering
  \footnotesize
  \begin{tabular}{|l|c|c||c|c|c|c|c|c|c|}
  \hline
  Method & PP & H $\times$ W & \cellcolor{col1}Abs Rel & \cellcolor{col1}Sq Rel & \cellcolor{col1}RMSE  & \cellcolor{col1}RMSE log & \cellcolor{col2}${\scriptstyle \delta < 1.25}$ & \cellcolor{col2}${\scriptstyle \delta < 1.25^{2}}$ & \cellcolor{col2}${\scriptstyle \delta < 1.25^{3}}$\\
  \hline 

% $L_{ps}$ Random SGM & \xmark &  $192 \times 640$ &
% 0.112 &	0.940 &	4.900 &	\textbf{0.195} &	0.870 &	\textbf{0.957} & \textbf{0.981} \\

% $L_{ps}$ Fused SGM & \xmark &  $192 \times 640$ &
% 0.112 &	0.912 &	4.876 &	0.199 &	0.869 &	0.954 &	0.978 \\

% $L_{sum}$ Fused SGM & \xmark  &  $192 \times 640$ &
% 0.111 &	0.889 &	4.833 &	0.196 &	0.869 &	0.956 &	\textbf{0.981} \\

% $L_{ps}$ Fused SGM $\rightarrow  L_{r}$ & \xmark &  $192 \times 640$ &
% 0.110 &	0.953 &	4.981 &	0.205 &	0.866 &	0.951 &	0.977 \\

% Klodt \cite{klodt2018supervising} uncertainty & \xmark & $192 \times 640$ &
% 0.110  &   0.928 &    4.873 &    0.198 &   0.869 & 0.954 &   0.979 \\

% \textbf{Ours} & \xmark  &  $192 \times 640$ &
% \textbf{0.108} & \textbf{0.829} & \textbf{4.782} &  \textbf{0.195} & \textbf{0.871} & 0.956 & 0.980 \\

% \arrayrulecolor{gray}\hline
% \arrayrulecolor{black}

% % Improved OpenCV Baseline
% $L_{ps}$ Random SGM HR & \xmark  & $320 \times 1024$ &
% 0.105 &	0.842 &	4.640 &	0.189 &	0.881 &	0.961 &	0.982
% \\

% $L_{ps}$ Fused SGM HR & \xmark  &  $320 \times 1024$ &

% \\

% $L_{sum}$ Fused SGM HR & \xmark  &  $320 \times 1024$ &
% \\

% \textbf{Ours HR} & \xmark  & $320 \times 1024$ & 0.101 &   0.752 &  4.527 &  0.188 & 0.885 & 0.961 & 0.981 \\

% \arrayrulecolor{gray}
\hline
% \hline
\arrayrulecolor{black}

$l_{ps}$ Random SGM  & \cmark  &  $192 \times 640$ &
0.110 &	0.901 &	4.816 &	\textbf{0.193} &	0.871 &	\textbf{0.958} &	\textbf{0.981} \\

$l_{ps}$ Random SGM LR  & \cmark  &  $192 \times 640$ &
0.109 &	0.877 &	4.800 &	\textbf{0.193} &	0.870 &	\textbf{0.958} &	\textbf{0.981} \\

$l_{ps}$ Fused SGM  & \cmark  &  $192 \times 640$ &
0.109 &	0.850 &	4.741 &		\textbf{0.193} &	0.873 &	0.956 &	0.980\\

$l_{\text{sum}}$ Fused SGM  & \cmark  &  $192 \times 640$ &
0.108 &	0.841 &	4.754 &	0.194 &	0.871 &	0.957 &	0.980 \\

$l_{ps}$ Fused SGM $\rightarrow  l_{r}$ & \cmark &  $192 \times 640$ &
0.109 &	0.916 &	4.910 &	0.203 &	0.866 &	0.952 &	0.977 \\

Klodt \cite{klodt2018supervising} uncertainty & \cmark & $192 \times 640$ &
% 0.110  &   0.928 &    4.873 &    0.198 &   0.869 & 0.954 &   0.979
0.108  &    0.905  &    4.815  &    0.196  &    0.871  &    0.955  &    0.979 
\\

\textbf{Ours} & \cmark  &  $192 \times 640$ &
\textbf{0.106} & \textbf{0.780} &  \textbf{4.695} & \textbf{0.193} & \textbf{0.875} & 	\textbf{0.958} & 0.980 \\

% \arrayrulecolor{gray}\hline
% \arrayrulecolor{black}

% % Improved OpenCV Baseline
% $l_{ps}$ Fused SGM (update) HR  & \cmark  & $320 \times 1024$ &
% \\

% $l_{ps}$ Random SGM HR  & \cmark  &  $320 \times 1024$ &
% 0.104 &	0.812 &	4.570 &	0.187 &	0.883 &	0.962 &	0.982 \\

% $l_{\text{sum}}$ Fused SGM HR  & \cmark  &  $320 \times 1024$ &
% \\

% \textbf{Ours HR } & \cmark  & $320 \times 1024$ & 0.099 &   0.727 &  4.465 &  0.186 & 0.887 & 0.962 & 0.982 \\

\arrayrulecolor{gray}\hline
\arrayrulecolor{black}

  \end{tabular}
  \vspace{3pt}
  \caption{\textbf{Ours \vs Baselines.} Comparison of baselines evaluated on KITTI 2015 \cite{Geiger2012CVPR} using the Eigen split. All methods here were trained on stereo pairs only. % Top section processes each image in a single pass. Bottom section shows results after post-processing.
  \vspace{-10pt}
  }
  \label{tab:ablation}
\end{table*}

\section{Experiments}
Our validation consists of four sets of experiments, all exploring the task of training a CNN to predict depth from a single color image, using binocular stereo data instead of ground-truth labels. Depending on the experiment, we compare against known leading baselines that supplement, and pre- and post-process the input stereo pairs and output depths to various degrees. The four experiments are:
\begin{enumerate}
\itemsep0em
\item Section~\ref{sec:ValidDepthHints} illustrates that local minima exist when photometric reconstruction loss is used for self-supervision, and that Depth Hints can help. 
\item Section~\ref{sec:ValidDHLoss} reports ablation-type experiments on Depth Hints, showing the negative impact of using the same SGM-computed stereo depths in more traditional loss functions. %\dt{is it a negative impact though? we get sota with baselines!}
\item Section~\ref{sec:ValidDHboost} shows how Depth Hints usually help other modern self-supervised models.
\item Section~\ref{sec:ValidDepthFromColorTask} pits Depth Hints against other state of the art algorithms, grouped by preconditions.
\end{enumerate}

We run experiments on the KITTI dataset~\cite{Geiger2012CVPR} which consists of calibrated stereo video registered to LiDAR measurements of a city, captured from a moving car. The depth evaluation is done on the LiDAR pointcloud, and we report all seven of the standard metrics. See \cite{eigen2014depth} for evaluation details, but broadly, lower numbers are better in red columns, while higher numbers are better in blue columns. To enable direct comparison with recent works, we use the Eigen split of KITTI~\cite{eigen2014depth} and evaluate with Garg's crop~\cite{garg2016unsupervised}, using the standard cap of 80m \cite{godard2017unsupervised}. 
We note that there are potential evaluation issues with the KITTI ground-truth data due to a translational offset between the color camera used to record images and the LiDAR scanner.
In the supplementary material we also present some evaluations on the updated KITTI ground truth data provided by~\cite{uhrig2017sparse}.

% The Eigen split we use follows Zhou~\etal~\cite{zhou2017unsupervised}'s preprocessing and consists of $19,905$ stereo pairs ($39,810$ images) for training. \gb{optional if we need space}

%%%%%%%%%%%%%%%%%%%%%%%%%%%%%%%%%%%%%%%%%%%%%%%%%%%%%%
\subsection{Solution with Depth Hints} \label{sec:ValidDepthHints}
%%%%%%%%%%%%%%%%%%%%%%%%%%%%%%%%%%%%%%%%%%%%%%%%%%%%%%
%In this subsection we offer an analysis of the produced monocular depth estimation networks.

The experiment described in Figure~\ref{fig:motivation} is typical, showing that recent self-supervision approaches can get by without ground truth depths for most pixels, because a DSSIM+$L_1$ loss trains the CNN to regress reasonable depths. However, even seemingly distinct structures like a tree induce local minima that are plausible, and hard for the training process to escape. Supervised training with LiDAR data would yield an excellent photometric match, but in its absence, a Depth Hint can provide an alternative that our loss function (\ref{eq:ours}) incorporates in a gradual way: the hint isn't trusted explicitly, and as training progresses, the hint may be ignored. 

In experiments, the network initially makes use of Depth Hints for $85$\% of available pixels, dropping to $50$\% at the end of training.
%we report on the network predictions during training. As you can see, with depth hints, the network converges to the globally optimal depth value. However, without the depth hints, the network fails to escape a local minimum and learns to predict sub-optimal depth value.
% Figure~\ref{fig:depthhints} shows how the network learns to predict depth maps that are similar in places to Depth Hints, but have good reprojection scores and less noise.

%%%%%%%%%%%%%%%%%%%%%%%%%%%%%%%%%%%%%%%%%%%%%%%%%%%%%%
\subsection{Baseline Loss Functions} \label{sec:ValidDHLoss}
%%%%%%%%%%%%%%%%%%%%%%%%%%%%%%%%%%%%%%%%%%%%%%%%%%%%%%

Besides our proposed loss in (\ref{eq:ours}), there are various alternative strategies for incorporating Depth Hints in the objective function. Here we discuss such alternatives and compare them experimentally in Table~\ref{tab:ablation}.

First, we start with a simple baseline, where a neural network is trained to predict depth labels produced by an off-the-shelf stereo algorithm. This baseline is trained with loss
\begin{equation}
l_{ps}(d_i) = l_{s}^{\log L_1}(d_i, h_i),
\end{equation}
where ``$ps$'' indicates proxy-supervised losses. Here $h_i$ is estimated by the SGM algorithm. We train three baselines with this loss. The first uses depth maps generated on the fly with a random selection of hyperparameters (Random SGM) to avoid the influence of DSSIM+$L_1$ loss. \rev{}{The second baseline uses the same method, but with a left-right consistency check to reduce noise by invalidating pixels which have disagreeing depth values in the two views (Random SGM LR).} The last baseline uses the single Fused SGM depth maps from Section~\ref{sec:TrainingFromDHstereo} that give an indirect signal from the DSSIM+$L_1$ loss.

Another approach is to optimize the sum of self-supervised and supervised losses, so
\begin{equation}
l_{\text{sum}}(d_i) = l_{r}(d_i) + l_{s}^{\log L_1}(d_i, h_i).
\end{equation}
This baseline is similar to the additional supervision from SLAM found in \cite{klodt2018supervising, yang2018deep}. Similarly, Zhu~\etal~\cite{guidedflow17} add a supervised loss \cite{bailer2015flow} to solve for optical flow and Kuznietsov~\etal~\cite{kuznietsov2017semi} add a supervised loss for depth estimation from LiDAR. \rev{}{Concurrently proposed monoResMatch~\cite{Tosi_2019_CVPR} uses this method to incorporate a proxy-supervised signal, albeit using a reverse Huber loss \cite{laina2016deeper} as opposed to $\log L_1$.} The addition of supervised losses change the objective function that is being minimized; one could view the additional term as a form of regularization, constraining the network prediction to adhere to the proposed depth values. However, this strategy can struggle to contend with noise in the depths estimated by stereo algorithms. 

A different way of incorporating Depth Hints is to pre-train a network using $l_{ps}$ on the fused Depth Hints and  fine-tune using $l_{r}$. In Table~\ref{tab:ablation}, this method is denoted as ``$l_{ps}$ Fused SGM $\rightarrow  l_{r}$''. We train $l_{ps}$ for 10 epochs followed by $l_{r}$ for another 10 epochs with the original learning rate.
% This strategy is called "pre-train + fine-tune"

Since the fused SGM depths may be a noisy estimate of depth, we could enable our model to train from them more robustly by explicitly modeling uncertainty \cite{kendall2017uncertainties, klodt2018supervising}.
In these prior works, imperfections in the supervisory signal are modelled as part of the training loss; in addition to disparity, the network predicts a per-pixel data-dependent estimate of the residual error of the supervised loss.
For pixels where the network expects that it will not be able to accurately satisfy the main training loss, it can pay a `penalty' by predicting a higher residual error.
This method was exploited by Klodt and Vedaldi~\cite{klodt2018supervising} to make learning from potentially noisy SLAM depths and poses more robust.

Referring to Table \ref{tab:ablation}, we note the clear benefit of treating the Depth Hints as noisy and only incorporating their estimates when they are superior to the network prediction. Surprisingly, our various baselines are competitive when compared to state of the art methods in Table~\ref{tab:kitti_eigen}. For example, even ``$l_{ps}$ Fused SGM'' scores better than 3Net~\cite{poggi20183net} and SuperDepth~\cite{pillai2018superdepth}, and is highly competitive with Monodepth2 (S and MS)~\cite{monodepth2} on all metrics, albeit with pre-training.

% At each pixel, the main loss is attenuated by the variance prediction.

\begin{table*}[th]
  \centering
    % \begin{minipage}[l]{1.0\textwidth}
%   \resizebox{0.7\textwidth}{!}{
  \centering
  \footnotesize
  \begin{tabular}{|c l|c|c|c|c||c|c|c|c|c|c|c|}
  \hline
  Cit. & Method & PP & Data & Dataset &H $\times$ W & \cellcolor{col1}Abs Rel & \cellcolor{col1}Sq Rel & \cellcolor{col1}RMSE  & \cellcolor{col1}RMSE log & \cellcolor{col2}${\scriptstyle \delta < 1.25}$ & \cellcolor{col2}${\scriptstyle \delta < 1.25^{2}}$ & \cellcolor{col2}${\scriptstyle \delta < 1.25^{3}}$\\
  \hline 

\cite{kuznietsov2017semi} & Kuznietsov & \xmark & DS & K & 192 $\times$ 640 &
0.109 & \textbf{0.693} & \textbf{4.305} & \textbf{0.176} & 0.878 & \textbf{0.965} & \textbf{0.987}
\\

\cite{kuznietsov2017semi} & \cellcolor{lightblue}Kuznietsov & \xmark & DS & K &192 $\times$ 640 &
\textbf{0.108} & \textbf{0.693} & 4.312 & \textbf{0.176} & \textbf{0.879} &  \textbf{0.965} &  0.986 
\\

\arrayrulecolor{black}
%\hline
%\cite{godard2017unsupervised} & Monodepth & \xmark & S & K &192 $\times$ 640 &
%0.120 & 1.068 & 5.116 & 0.206 & 0.856 & 0.949 & 0.977  \\

%\cite{godard2017unsupervised}  & \cellcolor{lightblue}Monodepth & \xmark & S & K &192 $\times$ 640 &
%\textbf{0.110} & \textbf{0.841} & \textbf{4.789} & \textbf{0.194} & \textbf{0.867} & \textbf{0.955} & \textbf{0.980} \\

\hline

\cite{monodepth2} & Monodepth2 no pt& \xmark & S & K &192 $\times$ 640 & 0.129 & 1.102 & 5.440 & 0.232 & 0.829 & 0.933 & 0.969 \\

\cite{monodepth2} & \cellcolor{lightblue}Monodepth2 no pt& \xmark & S & K &192 $\times$ 640 & \textbf{0.127} & \textbf{1.039} & \textbf{5.239} & \textbf{0.219} & \textbf{0.835} & \textbf{0.942} & \textbf{0.974} \\

\hline

\cite{monodepth2} & Monodepth2 & \xmark & S & K &192 $\times$ 640 &
0.110 & 0.896 & 4.986 & 0.208 & 0.866 & 0.948 & 0.975 
\\

\cite{monodepth2} & \cellcolor{lightblue}Monodepth2 & \xmark & S & K &192 $\times$ 640 &
\textbf{0.109} & \textbf{0.845} &  \textbf{4.800} &\textbf{ 0.196} & \textbf{0.870} & \textbf{0.956} & \textbf{0.980} \\

\hline

 \cite{poggi20183net} & 3Net (Resnet18) & \xmark & S & K &192 $\times$ 640 & 
       \textbf{0.112} &   0.953 &    5.007 &    0.207 &    0.862 &    0.949 &   \textbf{0.976} \\

\cite{poggi20183net} & \cellcolor{lightblue}3Net (Resnet18)  & \xmark & S & K &192 $\times$ 640 & 
 \textbf{0.112} &    \textbf{0.929} &   \textbf{4.960} &    \textbf{0.204} &    \textbf{0.867} &    \textbf{0.951} &    \textbf{0.976} \\

\hline

\cite{godard2017unsupervised} & Monodepth & \xmark & S & K &192 $\times$ 640 &
0.111 & 0.912 & 4.977 & 0.205 & 0.863 & 0.950 & \textbf{0.977} 
\\

\cite{godard2017unsupervised} & \cellcolor{lightblue}Monodepth & \xmark & S & K &192 $\times$ 640 &
\textbf{0.109} & \textbf{0.862} &  \textbf{4.862} &\textbf{ 0.201} & \textbf{0.868} & \textbf{0.952} & \textbf{0.977} \\

\hline

\cite{monodepth2} & Monodepth2 & \xmark & MS & K &320 $\times$ 1024 & 
% 0.114 & 0.991 & 5.029 & 0.203 & 0.864 & 0.951 & 0.978
0.106 &   0.806 &    4.630 &    0.193 &   0.876 & 0.958 &   0.980
\\ 

\cite{monodepth2} & \cellcolor{lightblue}Monodepth2 & \xmark & MS & K & 320 $\times$ 1024 & 
\textbf{0.100} &    \textbf{0.728} &    \textbf{4.469} &    \textbf{0.185} &   \textbf{0.885} &   \textbf{0.962} &   \textbf{0.982} \\

\hline \hline

\cite{monodepth2} & Monodepth2 & \xmark & S & SF &352 $\times$ 640 &
0.340 & 6.176 & 5.938 & 0.449 & 0.639 & 0.852 & 0.923 
\\

\cite{monodepth2} & \cellcolor{lightblue}Monodepth2 & \xmark & S & SF &352 $\times$ 640 &
\textbf{0.219} & \textbf{1.157} &  \textbf{3.889} &\textbf{ 0.344} & \textbf{0.706} & \textbf{0.900} & \textbf{0.953} \\

\arrayrulecolor{black}\hline
  \end{tabular}
%   }\hfill
%   \raisebox{3pt}{
%   \begin{minipage}[c]{0.28\textwidth}
    \vspace{1pt}
  \caption{\textbf{Depth Hints with Existing Methods.} Comparison of our implementations of existing methods with and without Depth Hints. The data used to train/test is defined in the \textit{Dataset} column, whereby `K' is for KITTI 2015~\cite{Geiger2012CVPR} using the Eigen split, and `SF' is for the FlyingThings3D Sceneflow dataset \cite{MIFDB16}. Highlighted methods are augmented \colorbox{lightblue}{with Depth Hints}, and score better than their regular counterparts almost universally. \cite{kuznietsov2017semi} is an exception, possibly because it already uses LiDAR data. We also show results for \cite{monodepth2} without ImageNet \cite{imagenet} pretraining, denoted as `Monodepth2 no pt'. \textit{Data column} (data source used for training): D refers to methods that use depth supervision at training time, S is for self-supervised training on stereo images, and MS is for models trained with stereo video. 
  \vspace{-6pt}
  }
  \label{tab:existing}
%   \end{minipage}}
\end{table*}

%%%%%%%%%%%%%%%%%%%%%%%%%%%%%%%%%%%%%%%%%%%%%%%%%%%%%%
\subsection{Depth Hints for Existing Methods} \label{sec:ValidDHboost}
%%%%%%%%%%%%%%%%%%%%%%%%%%%%%%%%%%%%%%%%%%%%%%%%%%%%%%

Here we demonstrate the benefits of using Depth Hints to improve existing methods. As most existing methods do not provide training code, we have implemented a selection of them that are trained with self-supervised loss. Hence, we modify our loss functions to closely match the selected methods, while keeping our network architecture, image resolution, optimization parameters, and number of epochs consistent across experiments.

Table~\ref{tab:existing} shows quantitative results of existing methods that were augmented with Depth Hints. We see noticeable improvements in all methods which are trained using stereo (S) and stereo video (MS), demonstrating the effectiveness of incorporating Depth Hints. Additionally, we do not observe an improvement for the semi-supervised case~\cite{kuznietsov2017semi}, nor do the comparatively noisy Depth Hints hurt its results. Please see supplementary material for additional information regarding these implementations.

\rev{}{Finally, Depth Hints show substantial improvements when trained and evaluated on synthetic FlyingThings3D Sceneflow dataset~\cite{mayer2015large}. The improvements are significant due to many objects with thin structures present in the dataset. These results demonstrate that Depth Hints can improve monocular depth estimation in various domains.}

\begin{figure*}[t]
\centering
\resizebox{1.0\textwidth}{!}{
\begin{tabular}{r@{\hskip 1mm}c@{\hskip 1mm}c@{\hskip 1mm}c@{\hskip 1mm}c}
\footnotesize
\vspace{-3pt}
\raisebox{14pt}{Input} &
\includegraphics[width=0.2\textwidth]{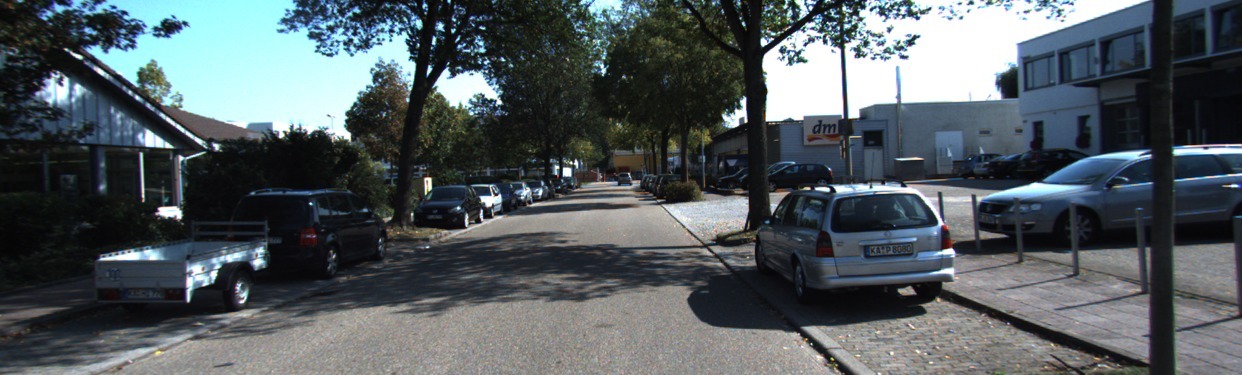} &
\includegraphics[width=0.2\textwidth]{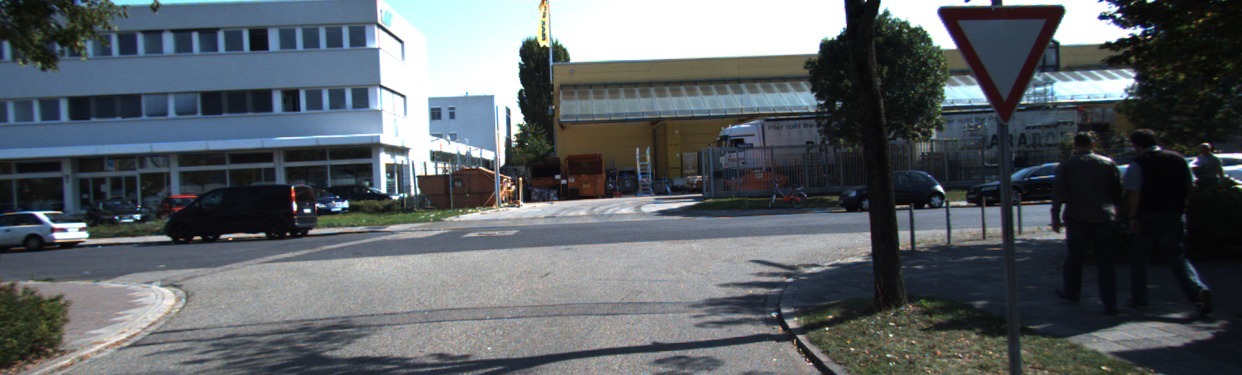} &
\includegraphics[width=0.2\textwidth]{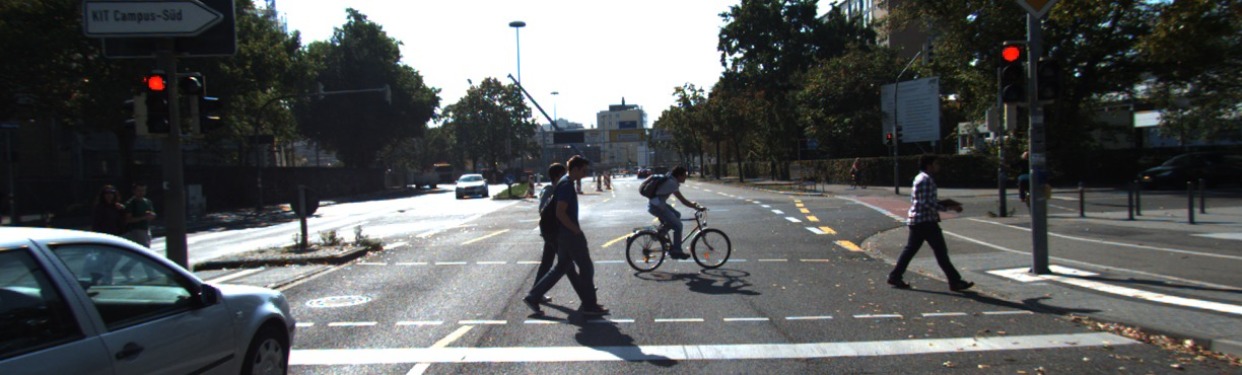} &
\includegraphics[width=0.2\textwidth]{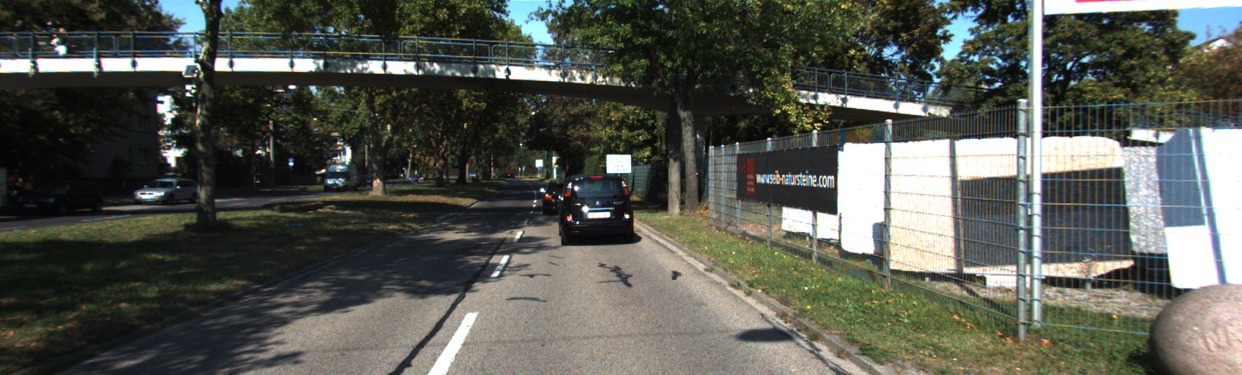} \\
% \vspace{-3pt}
% \footnotesize
% \raisebox{14pt}{Monodepth2~\cite{monodepth2}} &
% \includegraphics[width=0.2\textwidth]{figs/qual_results/030_mdp2_lowres.jpg} &
% \includegraphics[width=0.2\textwidth]{figs/qual_results/035_mdp2_lowres.jpg} &
% \includegraphics[width=0.2\textwidth]{figs/qual_results/545_mdp2_lowres.jpg} &
% \includegraphics[width=0.2\textwidth]{figs/qual_results/060_mdp2_lowres.jpg} \\
\vspace{-3pt}
\footnotesize
\raisebox{14pt}{3Net~\cite{poggi20183net}} &
\includegraphics[width=0.2\textwidth]{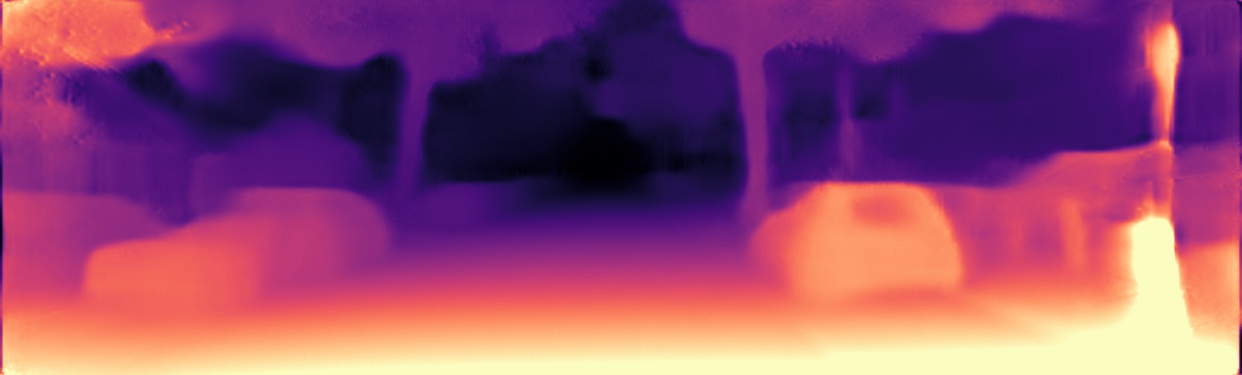} &
\includegraphics[width=0.2\textwidth]{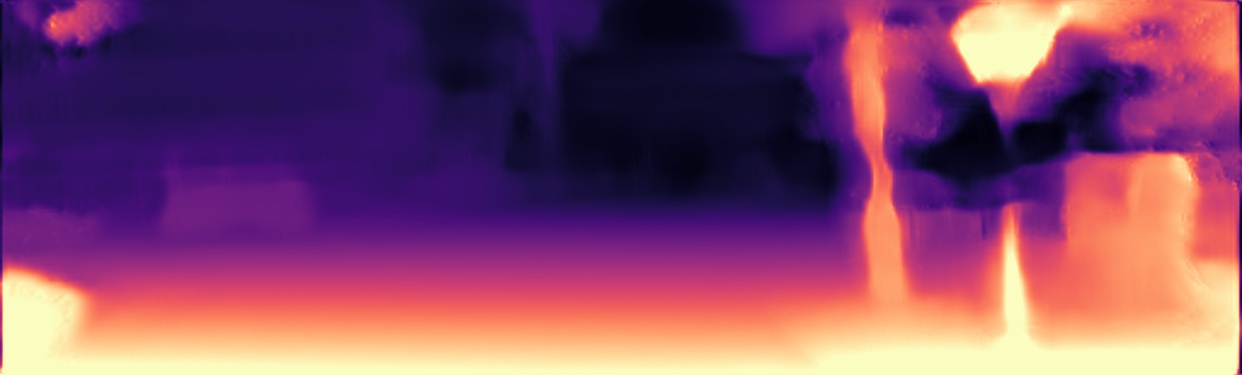} &
\includegraphics[width=0.2\textwidth]{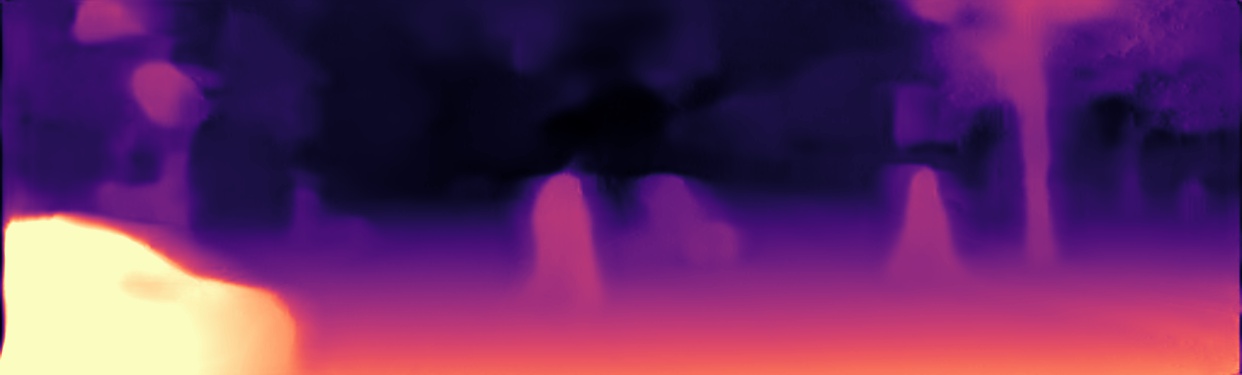} &
\includegraphics[width=0.2\textwidth]{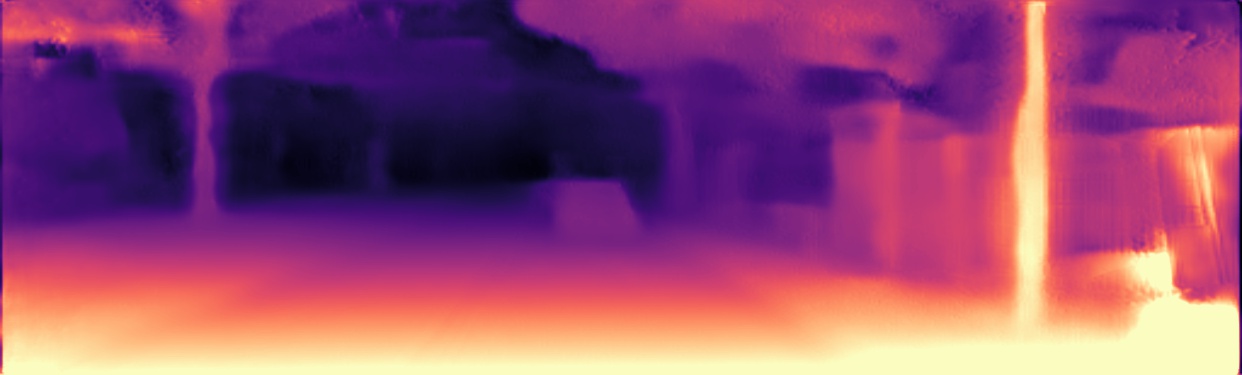} \\
\vspace{-3pt}
\footnotesize
\raisebox{14pt}{\textbf{Ours}} &
\includegraphics[width=0.2\textwidth]{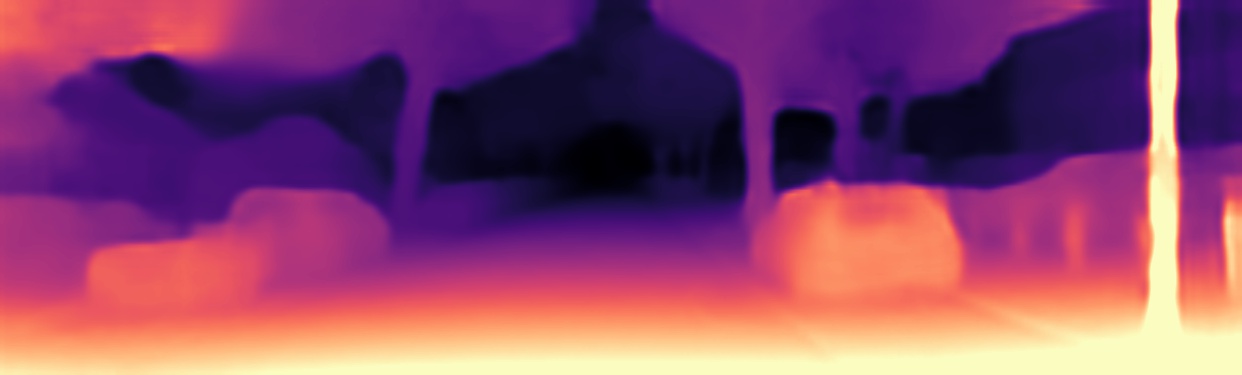} &
\includegraphics[width=0.2\textwidth]{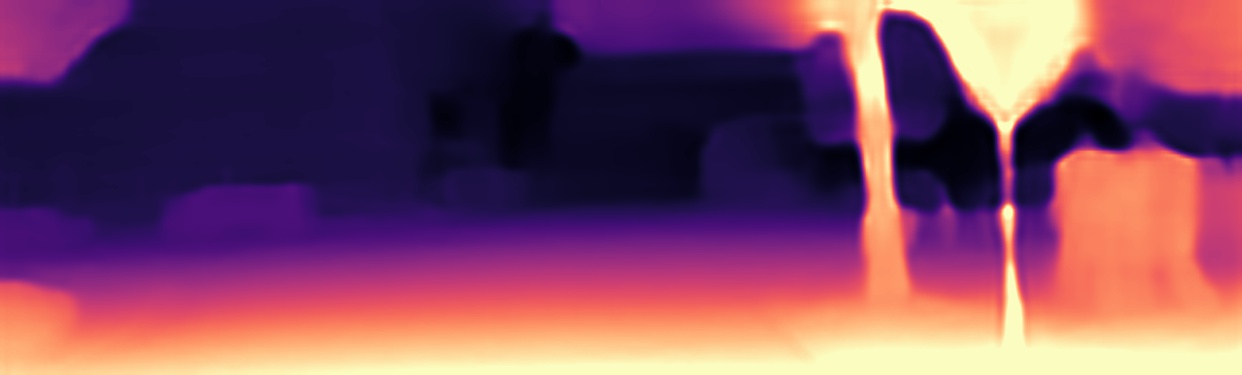} &
\includegraphics[width=0.2\textwidth]{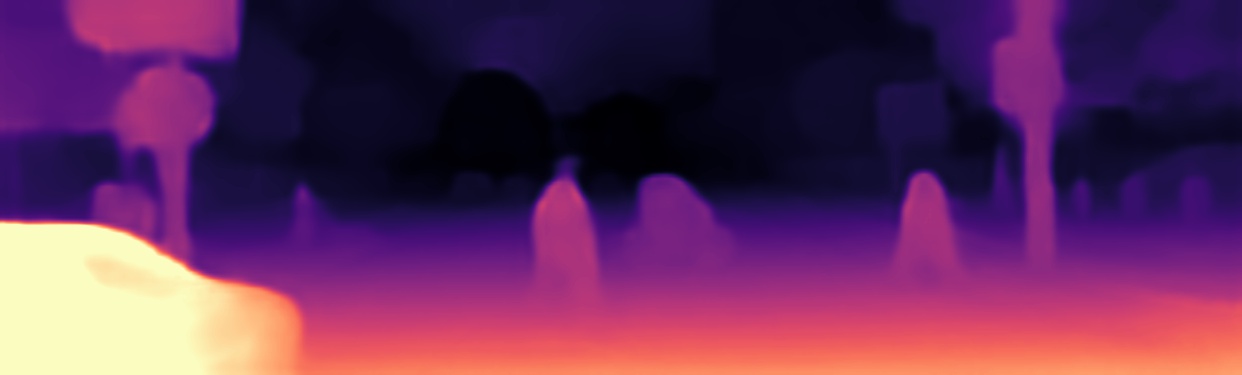} &
\includegraphics[width=0.2\textwidth]{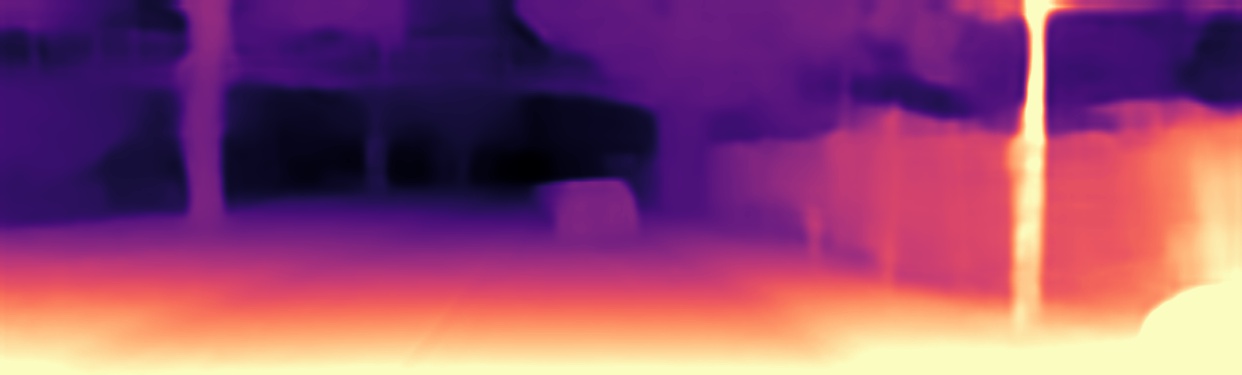} \\
\vspace{-3pt}
\footnotesize
\raisebox{14pt}{SuperDepth~\cite{pillai2018superdepth}} &
\includegraphics[width=0.2\textwidth]{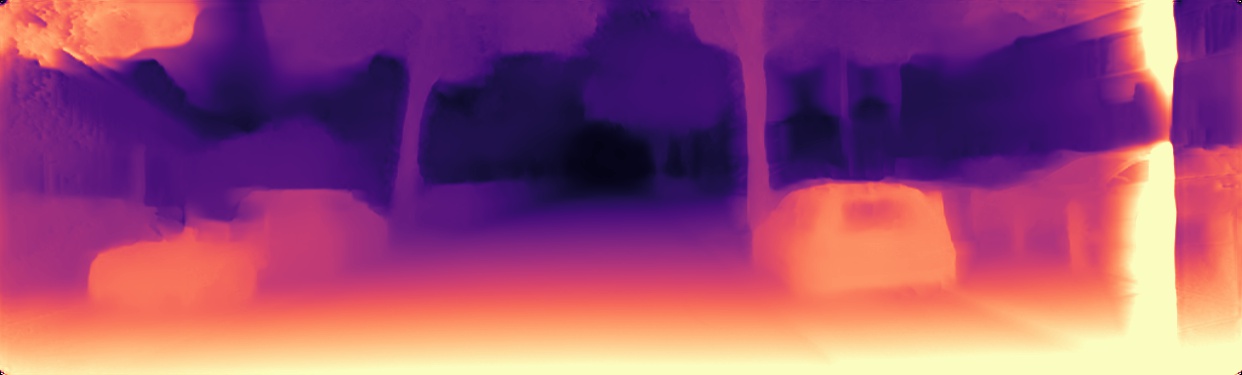} &
\includegraphics[width=0.2\textwidth]{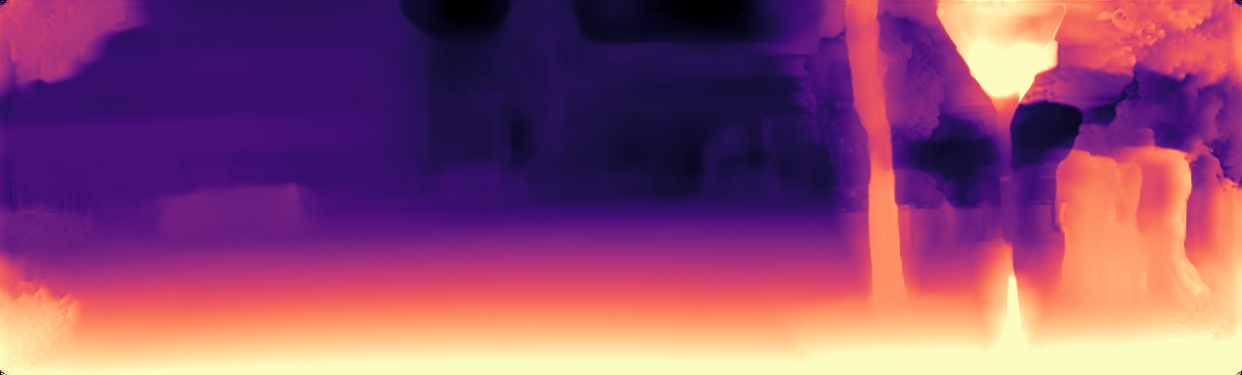} &
\includegraphics[width=0.2\textwidth]{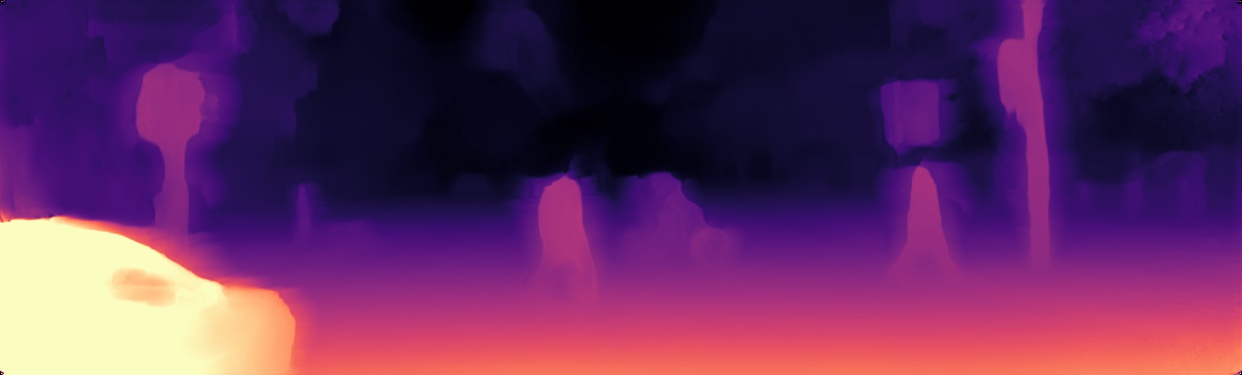} &
\includegraphics[width=0.2\textwidth]{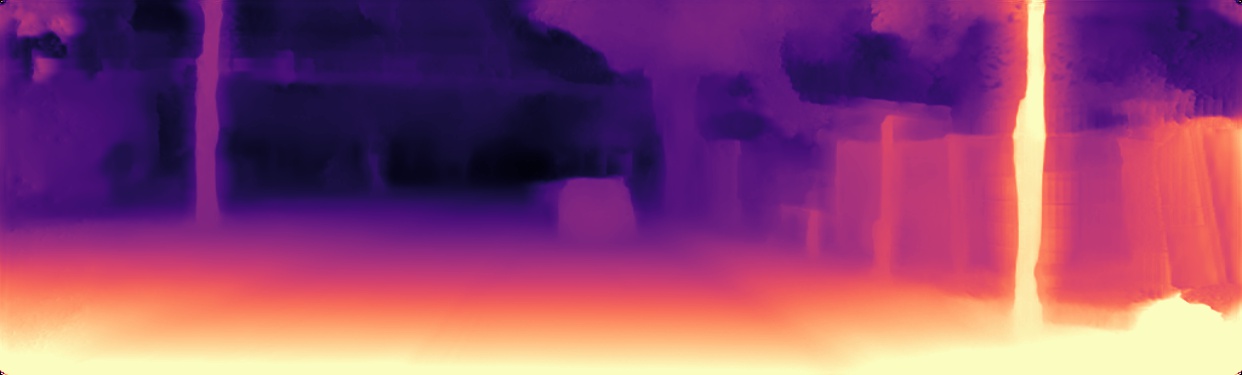} \\

\vspace{-3pt}
\footnotesize
\raisebox{14pt}{Monodepth2 HR~\cite{monodepth2}} &
\includegraphics[width=0.2\textwidth]{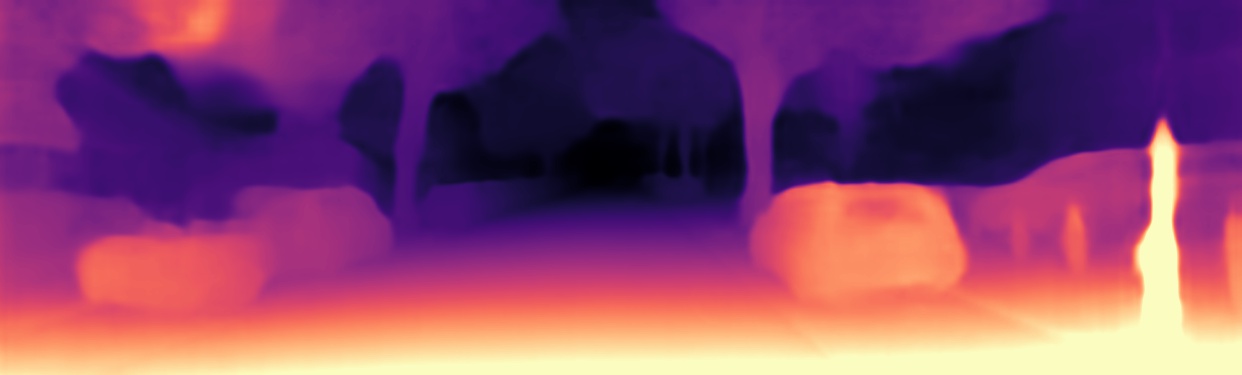} &
\includegraphics[width=0.2\textwidth]{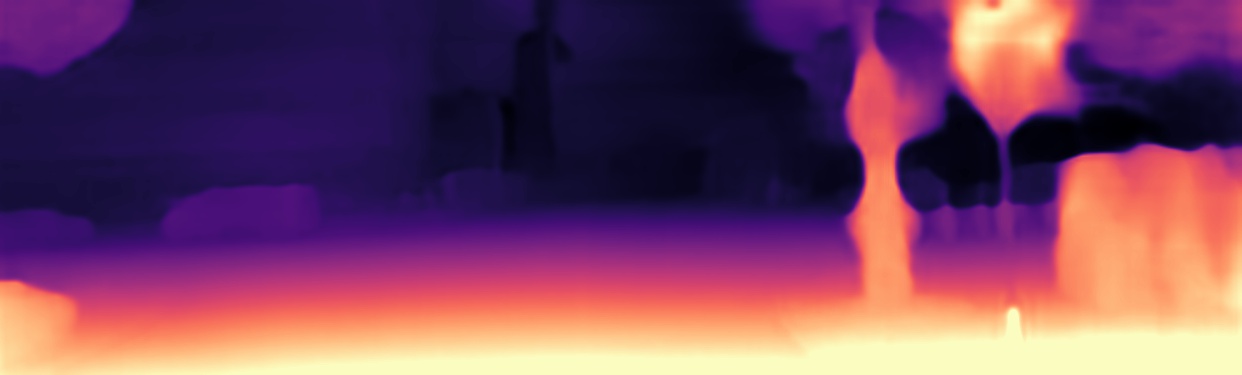} &
\includegraphics[width=0.2\textwidth]{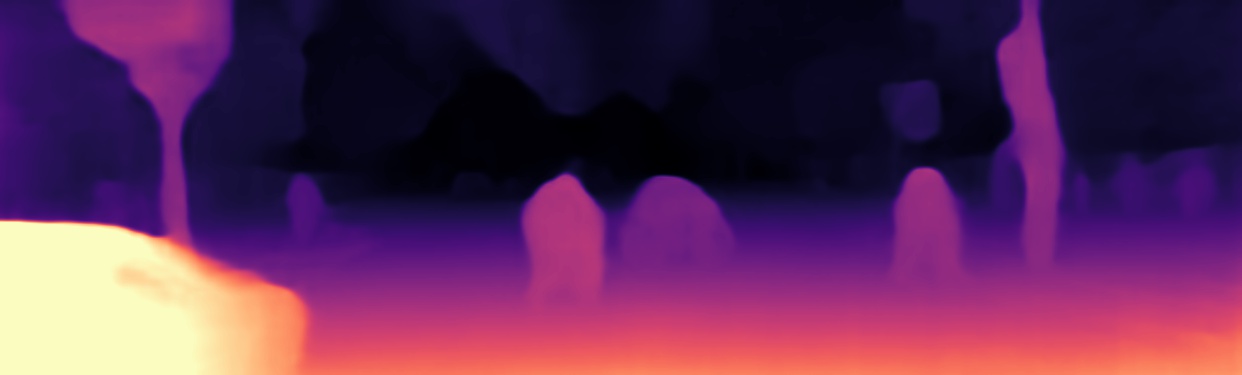} &
\includegraphics[width=0.2\textwidth]{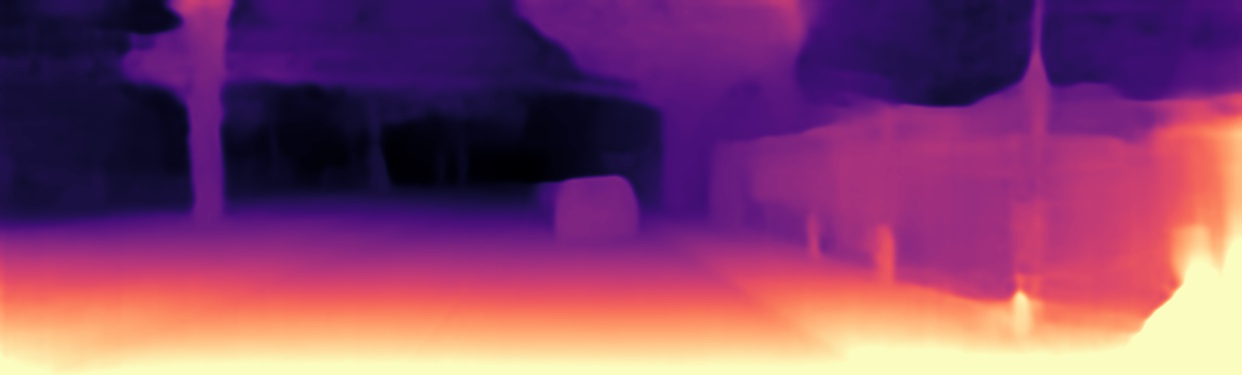} \\

\vspace{-3pt}
\footnotesize
\raisebox{14pt}{\textbf{Ours HR}} &
\includegraphics[width=0.2\textwidth]{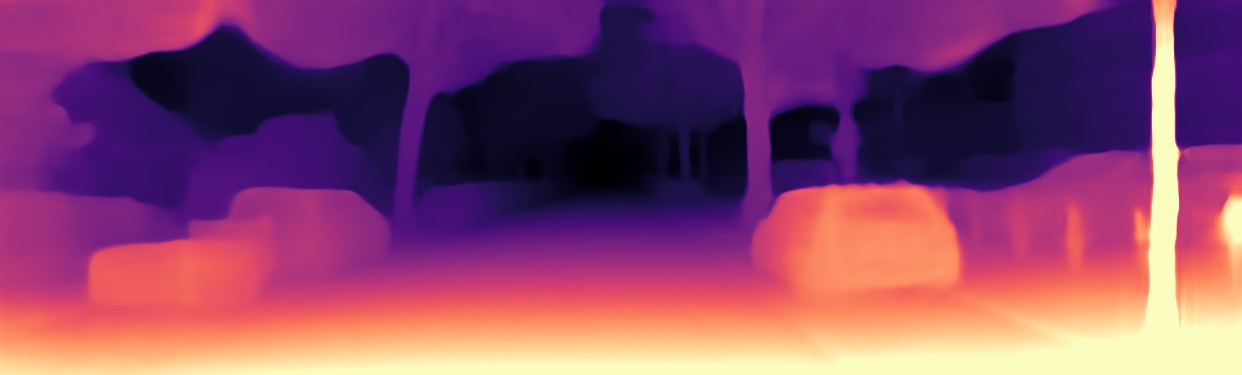} &
\includegraphics[width=0.2\textwidth]{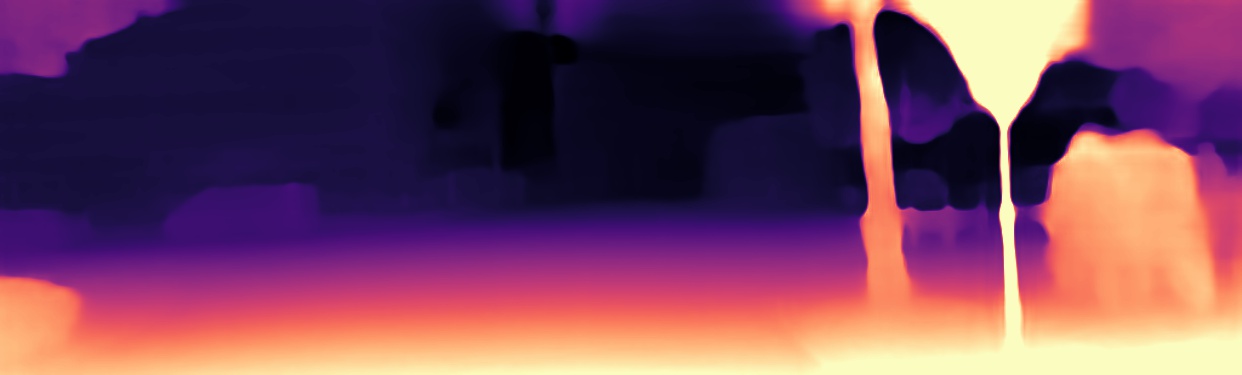} &
\includegraphics[width=0.2\textwidth]{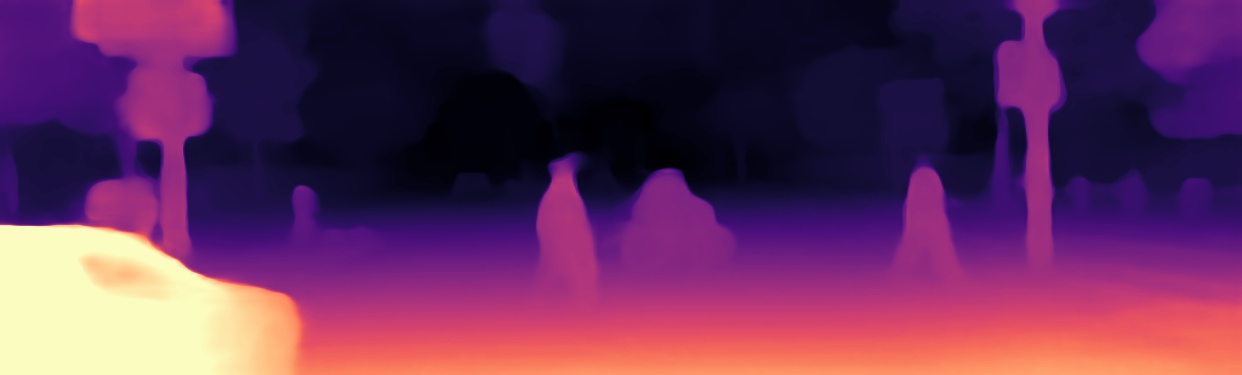} &
\includegraphics[width=0.2\textwidth]{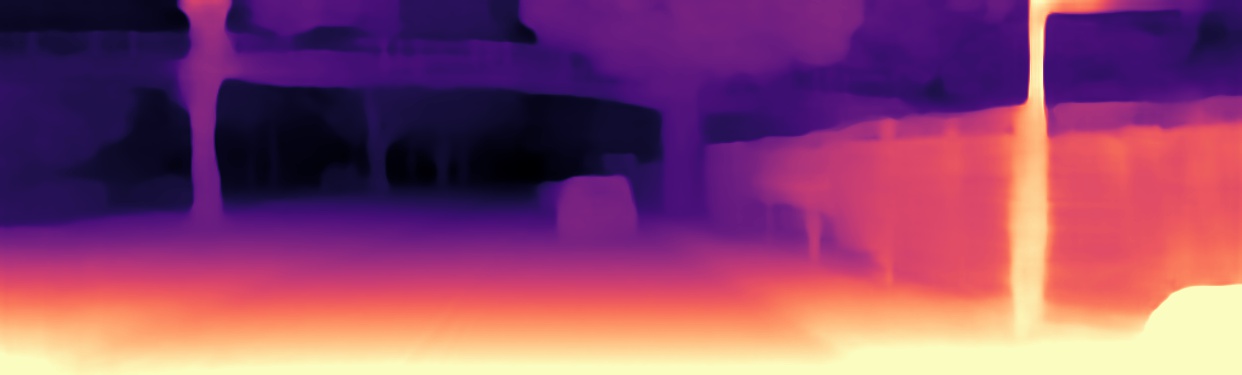} \\

\end{tabular}}
% \centering
% \begin{tabular}{*{4}{p{0.22\textwidth}}}
% \centering Test Set Image & \centering Godard~\etal~\cite{monodepth2} & \centering Ours & \centering Ours HR
% \end{tabular}
% }\end{minipage}
\caption{\textbf{Qualitative comparison with existing methods.} Top row: Four test set images. Each subsequent row: Depth maps generated by a stereo-only method. Notice how Ours and Ours HR capture thin structures such as traffic lights, traffic signs, lampposts, etc. %(Best viewed in color.)
}
\label{fig:qualitative}
\end{figure*}

%%%%%%%%%%%%%%%%%%%%%%%%%%%%%%%%%%%%%%%%%%%%%%%%%%%%%%
\section{Depth From Color Tournament} \label{sec:ValidDepthFromColorTask}
%%%%%%%%%%%%%%%%%%%%%%%%%%%%%%%%%%%%%%%%%%%%%%%%%%%%%%
Although it only represents one application domain, the KITTI dataset has been established as the dominant benchmark for measuring the accuracy of depth inferred from color. Broadly, our Depth Hints approach produces better looking results (see Figure~\ref{fig:qualitative}) and scores indicating that we are the new state of the art across three major competition ``categories.'' Please see Table~\ref{tab:kitti_eigen}. Of course there are more or less flattering ways to cluster the competition, so we present ``Our'' method in multiple forms, for better compatibility within each category. In doing so, we show that Depth Hints are useful across multiple settings (stereo \vs mono+stereo, low \vs high resolution, with/without pretraining),  making the difference between first and second place. %\jw{do we need to define S and MS explicitly again here? Its done in pervious section, but could write "...multiple settings stereo (S) vs stereo video (MS)" }%with/without post processing), 

%Figure~\ref{fig:qualitative} offers qualitative comparison of predicted depth maps for some of the test images of KITTI dataset.

% In this section, we compare the performance of our algorithm with state-of-the-art methods.

%Table~\ref{tab:kitti_eigen} shows our results on KITTI dataset.

Rows in Table~\ref{tab:kitti_eigen} are color-coded by category, with the winning score for each of seven measures marked in \textbf{bold}.

\begin{table*}[th]
  \centering
    % \begin{minipage}[l]{1.0\textwidth}
%   \resizebox{0.7\textwidth}{!}{
  \centering
  \footnotesize
  \resizebox{1.0\textwidth}{!}{
  \begin{tabular}{|c l|c|c|c||c|c|c|c|c|c|c|}
  \hline
  Cit. & Method & PP & Data & H $\times$ W & \cellcolor{col1}{\scriptsize Abs Rel} & \cellcolor{col1}{\scriptsize Sq Rel} & \cellcolor{col1}{\scriptsize RMSE}  & \cellcolor{col1}{\scriptsize RMSE log} & \cellcolor{col2}${\scriptstyle \delta < 1.25}$ & \cellcolor{col2}${\scriptstyle \delta < 1.25^{2}}$ & \cellcolor{col2}${\scriptstyle \delta < 1.25^{3}}$\\
  \hline 

% \cite{eigen2014depth} & Eigen & D & $172 \times 576$ & 
% 0.203 & 1.548 & 6.307 & 0.282 & 0.702 & 0.890 & 0.890 \\

% \cite{liu2015learning} & Liu & D &  &
% 0.201 & 1.584 & 6.471 & 0.273 & 0.680 & 0.898 & 0.967 \\

% \cite{klodt2018supervising} & Klodt & D*M &    &
% 0.166 & 1.490 & 5.998 & - &  0.778 & 0.919 & 0.966\\

% \cite{gandepth2018} & AdaDepth & D* &&  0.167 & 1.257 & 5.578 & 0.237 & 0.771 & 0.922 & 0.971\\

% \cite{zou2018df} & DF-Net & O*M & $160 \times 576$ & 0.150 & 1.124 & 5.507 & 0.223 & 0.806 & 0.933 & 0.973\\

\cite{kuznietsov2017semi} & \cellcolor{dsectioncolor} Kuznietsov &   & DS & 187 $\times$ 621 & 0.113 & 0.741 & 4.621 & 0.189 & 0.862 & 0.960 & 0.986\\

\cite{fu2018deep} & \cellcolor{dsectioncolor} DORN &   & D & 385 $\times$ 513 crop &
\textbf{0.072}&  \textbf{0.307} & \textbf{2.727} & \textbf{0.120} & \textbf{0.932} & \textbf{0.984} & \textbf{0.994}\\ 

\hline

\cite{yang2018deep} & \cellcolor{dsectioncolor} DVSO SimpleNet & \cmark   & D$^\dagger$S & 256 $\times$ 512 &  
0.107 & 0.852 & 4.785 & 0.199 & 0.866 & 0.950 & 0.978\\

\cite{yang2018deep} & \cellcolor{dsectioncolor} DVSO &  \cmark  & D$^\dagger$S & 256 $\times$ 512 &  
0.097 & 0.734 & 4.442 & 0.187 & 0.888 & 0.958 & 0.980\\

\hline

\cite{guo2018learning} & \cellcolor{dsectioncolor} Guo StereoUnsupFT $\rightarrow$ Mono pt &   & D*S & 256 $\times$ 512 &
0.099 & 0.745 & 4.424 & 0.182 & 0.884 & 0.963 & 0.983
\\

\cite{singlestereo2018} & \cellcolor{dsectioncolor} SVSM w/o finetuning &   & D*S & 192 $\times$ 640 crop &
0.102 & 0.700 & 4.681 & 0.200 & 0.872 & 0.954 & 0.978
\\

\cite{guo2018learning} & \cellcolor{dsectioncolor} Guo StereoSupFTAll $\rightarrow$ Mono pt &   & D*DS & 256 $\times$ 512 &
0.097 & 0.653 & \textbf{4.170} & \textbf{0.170} & 0.889 & \textbf{0.967} & \textbf{0.986}
\\

\cite{singlestereo2018} & \cellcolor{dsectioncolor} SVSM finetuned &   & D*DS & 192 $\times$ 640 crop &  \textbf{0.094} & \textbf{0.626} &4.252 & 0.177 & \textbf{0.891} & 0.965 & 0.984\\

% \arrayrulecolor{black}\hline

% \cite{zhou2017unsupervised} & Zhou \textdagger &   & M & 128 $\times$ 416$ & 
% 0.183 & 1.595 & 6.709 & 0.270 & 0.734 & 0.902 & 0.959\\

% \cite{yang2017unsupervised} & Yang & ? & M & 128 $\times$ 416 & 0.182 & 1.481  & 6.501  & 0.267  & 0.725  & 0.906  & 0.963\\

% \cite{mahjourian2018unsupervised} & Mahjourian &   & M & 128 $\times$ 416  & 
% 0.163 & 1.240 & 6.220 & 0.250 & 0.762 & 0.916 & 0.968\\

% \cite{geonet2018} & GeoNet &   & M  &  128 $\times$ 416  & 
% 0.155 & 1.296 & 5.857 & 0.233 & 0.793 & 0.931 & 0.973\\

% \cite{wang2017learning} & DDVO &   & M  &  128 $\times$ 416  & 
% 0.151 & 1.257 & 5.583 & 0.228 & 0.810 & 0.936 & 0.974\\ 

% \cite{ranjan2018adversarial} & Ranjan &   & M & 128 $ \times$ 416  & 
% 0.148 & 1.149 & 5.464 & 0.226 & 0.815 & 0.935 & 0.973\\

% \cite{luo2018every} & EPC++ &   & M &  256$ \times$ 832 & 
% 0.141 & 1.029 & 5.350 & 0.216 & 0.816 & 0.941 & 0.976\\

% \cite{monodepth2} & Godard &   & M &  192$ \times$ 640 & 
% 0.129 & 1.112 & 5.180 & 0.205 & 0.851 & 0.952 & 0.978 \\ 

% \textbf{Ours} & \\

\arrayrulecolor{black}
\hline
\hline

% \cite{garg2016unsupervised} & \cellcolor{ssectioncolor} Garg $\ddagger$ & \xmark & S  & 188 $\times$ 620 &
% 0.152 & 1.226 & 5.849 & 0.246 & 0.784 & 0.921 & 0.967\\

\cite{godard2017unsupervised} & \cellcolor{ssectioncolor} Monodepth & \cmark & S & 256 $\times$ 512 &
% 0.133 & 1.142 & 5.533 & 0.230 & 0.830 & 0.936 & 0.970
0.138 &   1.186 &   5.650 &   0.234 &   0.813 &   0.930 &   0.969
\\

\cite{mehta2018structured} & \cellcolor{ssectioncolor} StrAT &  & S  & 256 $\times$ 512 &
0.128 & 1.019 & 5.403 & 0.227 & 0.827 & 0.935 & 0.971\\

% 
% \cite{poggi20183net} & \cellcolor{ssectioncolor} 3Net (VGG) pp &   & S & 256 $\times$ 512 & 
% 0.114 & 1.088 & 5.756 & 0.203 & 0.848 & 0.944 & 0.979 \\

\cite{monodepth2} & \cellcolor{ssectioncolor} Monodepth2 (w/o pretraining) & \cmark & S & 192 $\times$ 640 &
0.128 & 1.089 & 5.385 & 0.229 & 0.832 & 0.934 & 0.969  \\  

\cite{poggi20183net} & \cellcolor{ssectioncolor} 3Net (Resnet50) & \cmark & S & 256 $\times$ 512 & 
0.126 & 0.961 & 5.205 & 0.220 & 0.835 & 0.941 & 0.974 \\

& \cellcolor{ssectioncolor} \textbf{Ours Resnet50 w/o pretraining} & \cmark & S & 192 $\times$ 640 &  0.118 &   0.941 & 5.055 & 0.210 & 0.850 & 0.948 & 0.976 \\

\cite{monodepth2} & \cellcolor{ssectioncolor} Monodepth2 & \cmark & S & 192 $\times$ 640 &
0.108 & 0.842 & 4.891 & 0.207 & 0.866 & 0.949 & 0.976  \\  

% 
% \cite{klodt2018supervising} & \cellcolor{ssectioncolor} Klodt (our implementation*) & & S?? & 192 $\times$ 640 &
% 0.110  &   0.928 &    4.873 &    0.198 &   0.869 & 0.954 &   0.979 \\

& \cellcolor{ssectioncolor} \textbf{Ours} & \cmark & S & 192 $\times$ 640 &
0.106 &   0.780 &  4.695 &  0.193 & 0.875 & 0.958 & 0.980\\

& \cellcolor{ssectioncolor} \textbf{Ours Resnet50} & \cmark & S & 192 $\times$ 640 &
\textbf{0.102} &   \textbf{0.762} &  \textbf{4.602} &  \textbf{0.189} & \textbf{0.880} & \textbf{0.960} & \textbf{0.981}\\

\hline

\cite{pillai2018superdepth} & \cellcolor{hrsectioncolor} SuperDepth & \cmark & S & 384 $
\times$ 1024 & 0.112 & 0.875 & 4.958 & 0.207 & 0.852 & 0.947 & 0.977\\

& \cellcolor{hrsectioncolor} \textbf{Ours HR Resnet50 w/o pretraining} & \cmark & S & 320 $\times$ 1024 &  0.112 &   0.857 & 4.807 & 0.203 & 0.861 & 0.952 & 0.978 \\

 \cite{Tosi_2019_CVPR} & \cellcolor{hrsectioncolor} monoResMatch & \cmark & S & 256 $\times$ 512 crop &
 0.111 & 0.867 & 4.714 & 0.199 & 0.864 & 0.954 & 0.979 \\

\cite{monodepth2} & \cellcolor{hrsectioncolor} Monodepth2 & \cmark & S & 320 $\times$ 1024 &
0.105 & 0.822 & 4.692 & 0.199 & 0.876 & 0.954 & 0.977  \\  

& \cellcolor{hrsectioncolor} \textbf{Ours HR} & \cmark & S & 320 $\times$ 1024 & 0.099 &   0.723 &  4.445 &  0.187 & 0.886 & \textbf{0.962} & \textbf{0.981} \\

& \cellcolor{hrsectioncolor} \textbf{Ours HR Resnet50} & \cmark & S & 320 $\times$ 1024 & \textbf{0.096} &   \textbf{0.710} &  \textbf{4.393} &  \textbf{0.185} & \textbf{0.890} & \textbf{0.962} & \textbf{0.981} \\

%\hline

%& \cellcolor{cropsectioncolor} \textbf{Ours crop w/o pretraining} & \cmark & S & 320 $\times$ 512 crop & 0.115 &   0.873 &  4.847 &  0.205 & 0.857 & 0.950 & 0.977 \\

%& \cellcolor{cropsectioncolor} \textbf{Ours crop} & \cmark & S & 320 $\times$ 512 crop & 0.100 &   0.739 &  4.511 &  0.190 & 0.883 & 0.960 & 0.981 \\

\arrayrulecolor{black}
\hline
\hline

%\cite{li2017undeepvo} & \cellcolor{mssectioncolor} UnDeepVO &   & MS  & 128 $\times$ 416 &  0.183 & 1.730 & 6.57 & 0.268 & - & - & -\\

\cite{zhanst2018} & \cellcolor{mssectioncolor} Zhan & \xmark & MS  & 160 $\times$ 608 &  0.135 & 1.132 & 5.585 & 0.229 & 0.820 & 0.933 & 0.971\\ 

\cite{luo2018every} & \cellcolor{mssectioncolor} EPC++ &   & MS & 256 $\times$ 832 & 0.128 & 0.935 & 5.011 & 0.209 & 0.831 & 0.945 & 0.979 \\

\cite{monodepth2} & \cellcolor{mssectioncolor} Monodepth2 & \cmark & MS & 192 $\times$ 640 & 
\textbf{0.104} & 0.786 & 4.687 & 0.194 & \textbf{0.876} & 0.958 & 0.980 \\ 

& \cellcolor{mssectioncolor} \textbf{Ours} & \cmark & MS & 192 $\times$ 640 &
0.105 &    \textbf{0.769} & \textbf{4.627} & \textbf{0.189} &  0.875 &    \textbf{0.959} &   \textbf{0.982} \\
% & \textbf{Ours} 3 frames full mask & \cmark & MS  & $192 \times 640$ &    
% 0.106 &    0.807 &    4.694 &    0.191 &    0.876 &    0.958 &    0.981\\

\arrayrulecolor{black}
\hline

\cite{monodepth2} & \cellcolor{mssectioncolor} Monodepth2 & \cmark & MS & 320 $\times$ 1024 & 0.104 &    0.775 &   4.562 &    0.191 &   0.878 &   0.959 &   0.981 \\

& \cellcolor{mssectioncolor} \textbf{Ours HR} & \cmark & MS & 320 $\times$ 1024 & \textbf{0.098} &    \textbf{0.702} &    \textbf{4.398} &    \textbf{0.183} &   \textbf{0.887} &   \textbf{0.963} &   \textbf{0.983} \\

\arrayrulecolor{black}\hline
  \end{tabular}}
%   }\hfill
%   \raisebox{3pt}{
%   \begin{minipage}[c]{0.28\textwidth}
    \vspace{3pt}
  \caption{\textbf{Quantitative results.} Adjusting our model slightly, we compare it to the top performers in three different categories on KITTI 2015~\cite{Geiger2012CVPR}, using the Eigen split. % \newline
  \textit{Data column} (data source used for training): D refers to methods that use KITTI depth supervision at training time, D* use auxiliary depth supervision from synthetic data, D$^\dagger$ use auxiliary depth supervision from SLAM, %P$^\dagger$ use poses estimated with ORB-SLAM2~\cite{orbslam2}, 
  % M indicates self-supervised algorithms trained on monocular videos, 
  S is for self-supervised training on stereo images, MS is for models trained with both M (forward and backward frames) and S data.
  % $\ddagger$ ~indicates newer results from the respective online implementations.
  % *~indicates reimplementation of their loss with our architecture.
%   \gb{If needed, can label First, Second, Third place winners within a category using ``badges'' $\circone$, $\circtwo$, $\circthree$}
  }
  \vspace{-6pt}
  \label{tab:kitti_eigen}
%   \end{minipage}}
\end{table*}

\textbf{\colorbox{ssectioncolor}{Low-res Stereo}} is the classic category, with the longest history of competitors (we show the highest scorers). Our full method (``Ours Resnet50'') wins decisively on every metric. One could argue about two ``outside'' advantages: we pre-train on Imagenet and our SGM step gets the benefit of a time-tested heuristic. Our ablation experiments in Sec~\ref{sec:ValidDHLoss} show the difference between using SGM naively and incorporating its output as a Depth Hint. For completeness, we present results for our method with no pretraining (``Ours Resnet50 w/o pretraining''). When we compare this to the highest scoring non pretrained network 3Net~\cite{poggi20183net}, we show better scores in all seven metrics. 

%Since 3Net uses a larger encoder and trains at a different resolution, we show a like-for-like comparison in Table~\ref{tab:existing} demonstrating the positive impact of Depth Hints. Additionally, without pretraining our method is still competitive with Monodepth2~\cite{monodepth2} which is pretrained.

%\dt{Ours vs SOTA. We beat with pre-trained nets everyone.
%We do worse without pretraining, but still competitive. Closest are 3Net which do not pretrain, and they are better on only one metric, however they have a bigger encoder, although our resolutions are different.}

\rev{}{\textbf{\colorbox{hrsectioncolor}{High-res}} allows for processing of larger inputs. Again our method (``Ours HR Resnet50'') shows a considerable improvement over existing methods in all metrics. Similar to before, we also show results for our method without pretraining (``Ours HR Resnet50 w/o pretraining''). Our non-pretrained model beats SuperDepth \cite{pillai2018superdepth} in six out of seven metrics (tied in one), and compares favourably to the concurrent work monoResMatch~\cite{Tosi_2019_CVPR}, which makes use of a significantly more complex network compared to our encoder-decoder architecture.}

%\textbf{\colorbox{hrsectioncolor}{High-res} and \colorbox{cropsectioncolor}{Crop} Stereo} are two sub-categories that allow processing of larger inputs. ``Ours HR'' compares favorably with SuperDepth~\cite{pillai2018superdepth} across all metrics and by a good margin. Superdepth does not pre-train, but it has sophisticated upsampling in the decoder, and they beat our no-pretrain version on only two metrics, having trained for $200$ epochs as opposed to our $30$.

\rev{}{\textbf{\colorbox{mssectioncolor}{Stereo Video MS}} could theoretically be the category with the strongest scores, because each self-supervised algorithm has access to time series movies (M) in stereo (S), with the opportunity to match occluded regions by searching elsewhere in time. Interestingly, in this category we see smaller improvements by using our approach over \cite{monodepth2} for lower resolution (``Ours''), but observe a substantial boost in the high resolution case (``Ours HR'').}

%\textbf{\colorbox{mssectioncolor}{Stereo Video MS}} could theoretically be the category with the strongest scores, because each self-supervised algorithm has access to time series movies (M) in stereo (S), with the opportunity to match occluded regions by searching elsewhere in time. Our approach again wins decisively on all seven metrics, with Monodepth2 \cite{godard2017unsupervised} (also pre-trained) coming closest. We were a little surprised that our scores here were only slightly better in places or slightly worse than our own results on \colorbox{ssectioncolor}{Low-res Stereo}. Speculating that the added load of learning to estimate poses was counteracting the access to richer data, we also did a control experiment. In the white category at the bottom of the table, we used poses from ORB-SLAM2~\cite{orbslam2}. Given those poses for the MS category, we do better, but only slightly. We hope that modeling moving objects with some form of Flow Hints in the future could give our MS algorithm an edge over the S version.
%\dt{We do slightly better with multiple frames on some metrics and slightly worse on others compared to our stereo. We suspect it is due to moving objects rather than estimated poses. As a control experiment, we use poses from ORB-SLAM2~\cite{orbslam2} and do better, but only slighty. If we have access to depth from stereo like DVSO~\cite{yang2018deep}, we do BLAH. However, adding modeling motion and adding something like flow-hints seems to be an exciting direction to pursue. }

Overall, we note that error metrics like SqRel and RMSE, which penalize large errors in a few pixels, benefit most from Depth Hints. %SqRel and RMSE are examples of these metrics, which measure the squared difference between  
Depth Hints help to recover thin structures and to more accurately delineate object boundaries (Figure~\ref{fig:qualitative}).
The AbsRel metric has smaller gains, since only a minority of pixels in each image are improved.

% \jw{do we want to keep the below about supervised? Might help for space if we cut it}
The \textbf{\colorbox{dsectioncolor}{Depth Supervised}} category is one we cannot compete in. The clear winner here is DORN~\cite{fu2018deep}, who avoid self-supervision entirely, training directly from LiDAR data. 
% Kuznietsov~\cite{kuznietsov2017semi}, an earlier work uses both LiDAR and stereo footage. 
SVSM~\cite{singlestereo2018} uses outside synthetic data, and LiDAR data for finetuning. DVSO~\cite{yang2018deep} obtains depth supervision through an excellent SLAM system, yielding LiDAR-like pointclouds, and combines them with self-supervision to achieve scores similar to ours in their ``SimpleNet'' model. However, their paper introduces an important enhancement that we lack, namely a depth refinement network. % \jw{what do we want to do about DORNs numbers? The current table is actually new ground truth not sparse - sparse results are much more in line with SVSM etc, but they only published new ground truth}

%\jw{resnet50 nopt trained for 25 epochs with LR step at 20, Ours MS HR trained for 9 epochs, step at 5, other all 20 epochs step at 5. Batchsize 12 for lowres, 6 for HR %and 4 for MS HR}

%%%%%%%%%%%%%%%%%%%%%%%%%%%%%%%%%%%%%%%%%%%%%%%%%%%%%%
\section{Conclusion}
%%%%%%%%%%%%%%%%%%%%%%%%%%%%%%%%%%%%%%%%%%%%%%%%%%%%%%
\vspace{-5pt}
We investigated current issues with reprojection losses in the self-supervised monocular depth estimation setting. Based on these observations, we introduced Depth Hints as a practical approach to help escape from local minima, and to guide the network toward a better overall solution. The depth proposals make for a strong baseline themselves, but our training mechanism reverts to the default reprojection loss when the proposals are unhelpful. 
Qualitatively, Depth Hints seem to help most with thin structures and sharp boundaries. Extensive experimentation supports this. Further, Depth Hints provide a boost when applied to existing self-supervision schemes. Combined with a common network architecture, without but preferably with pre-training, our Depth Hints model achieves the top-scores on the self-supervised KITTI Eigen benchmark by a significant margin. %, and shows that Depth Hints improved the quality of the depth prediction, and the analysis of the achieved solutions demonstrates that the solutions converge to a better global optimum. \dt{verification required}.

% We see potential for similar analysis of other photometric losses, such as normalized cross-correlation, or perceptual losses such as \cite{zhanst2018}.

\section*{Acknowledgements}

We would like to thank Aron Monszpart and Galen Han for helping to run our experiments, and our anonymous reviewers for their positive comments and helpful suggestions.

{\small
\bibliographystyle{ieee_fullname}
\bibliography{main}
}

\end{document}